\documentclass{l4dc2027}
\usepackage{times}

\usepackage{bm}
\usepackage{booktabs}
\usepackage{adjustbox}
\usepackage{enumitem}
\usepackage{placeins}
\usepackage{tikz}
\usepackage{longtable}
\usetikzlibrary{arrows.meta,positioning,calc}
\usepackage{hyperref}

\usepackage{amsmath,amsfonts,bm}

\def\eqref#1{equation~\ref{#1}}

\def\1{\bm{1}}

\DeclareMathAlphabet{\mathsfit}{\encodingdefault}{\sfdefault}{m}{sl}
\SetMathAlphabet{\mathsfit}{bold}{\encodingdefault}{\sfdefault}{bx}{n}

\newtheorem{rifttheorem}{Theorem}
\newtheorem{riftproposition}{Proposition}
\newtheorem{riftlemma}{Lemma}
\newtheorem{riftassumption}{Assumption}

\newcommand{\score}[2]{\ensuremath{#1\pm#2}}
\newcommand{\bestscore}[2]{\ensuremath{\bm{#1\pm#2}}}

\title[RIFT: Reward Improvement Field Transfer]{RIFT: Reward Improvement Field Transfer for Cross-Domain Offline RL}

\author{%
  \Name{Jinhee Won} \Email{jinhee@mit.edu}\\
  \Name{Mardavij Roozbehani} \Email{mardavij@mit.edu}\\
  \Name{Munther Dahleh} \Email{dahleh@mit.edu}\\
  \addr Massachusetts Institute of Technology, Cambridge, MA, USA
}

\begin{document}

\maketitle

\begin{abstract}
Learning from limited target data is a central challenge in cross-domain offline reinforcement learning. Abundant source data can help, but dynamics differences can make direct transition reuse unreliable. We propose Reward Improvement Field Transfer (RIFT), which instead learns how low- and high-reward source transitions differ. RIFT encodes source and target transitions in a shared latent space, pairs low- and high-reward source transitions using optimal transport, and fits an affine field to their displacements. It applies this field to low-reward target anchors and decodes the results to augment the target dataset. The approach builds on the hypothesis that related domains can share useful low-to-high reward structure despite different dynamics. We bound synthesis error under shared class geometry, identify conditions favoring target anchoring over source-high transplant, and derive a conditional connection to downstream return. Experiments on dense-reward locomotion tasks under gravity, morphology, and friction shifts show improved aggregate performance over existing baselines.
\end{abstract}


\section{Introduction}

Learning a good policy often requires more target-domain experience than is practical to collect. A robot may have abundant simulation data but little real-world data, or its dynamics may change with friction, payload, or body structure \citep{tobin2017domain,chebotar2019closing}. Cross-domain offline reinforcement learning (RL) studies how to use abundant source data and scarce target data without further interaction \citep{liu2022dara,lyu2025otdf}.

These datasets provide different kinds of information. Source data offer extensive experience, but dynamics mismatch makes direct reuse difficult: the same state and action can produce different next states, even under a shared task and reward-scoring rule \citep{peng2018sim}. Target data reflect the deployment dynamics, but may provide too little coverage to learn a good policy alone. The challenge is therefore to extract useful source information without assuming that source transitions accurately describe the target environment.

One possible source of information is how low- and high-reward transitions differ. This relationship may be easier to estimate from abundant source data than from scarce target data. For example, a locomotion dataset may contain transitions with similar forward progress but different control costs, suggesting ways to achieve comparable progress with less effort. A change in gravity or body structure may alter the exact motion needed, while some of this reward-related structure may remain useful. This motivates our central question: instead of transferring source transitions directly, can we transfer \emph{how behavior improves}?

\begin{figure}
    \centering
    \includegraphics[width=0.75\linewidth]{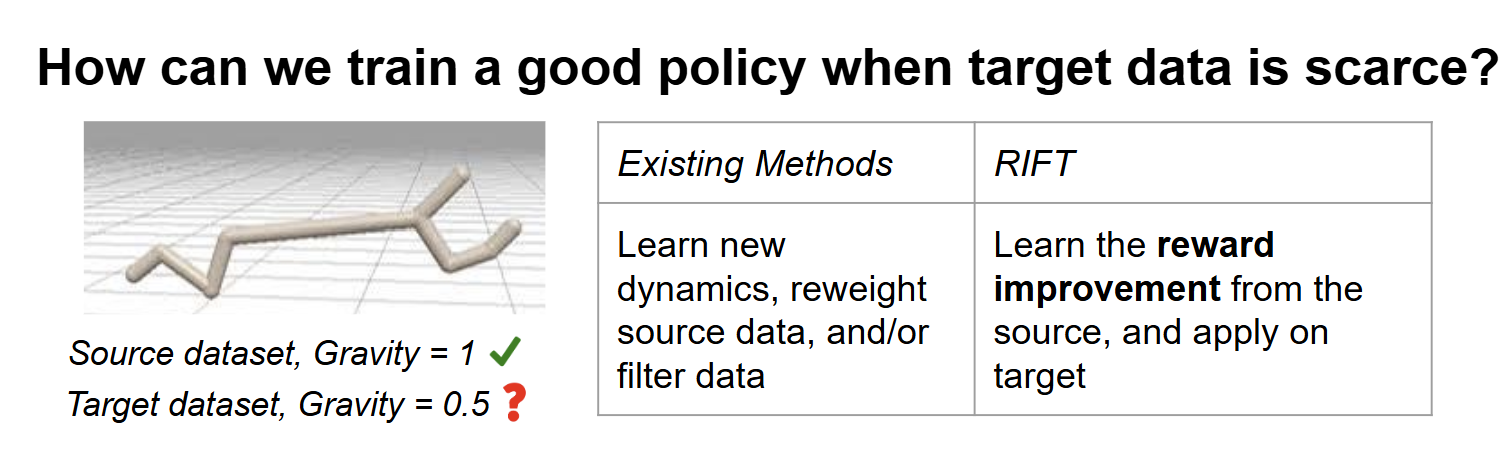}
    \caption{Motivation behind RIFT. A scarce dataset makes it difficult to learn in the target domain. How can we instead use existing structures from the source dataset to transfer what we learn?}
    \label{fig:motivation}
\end{figure}

We propose \textbf{R}eward \textbf{I}mprovement \textbf{F}ield \textbf{T}ransfer \textbf{(RIFT)}. RIFT maps source and target transitions into a shared latent space, matches low- and high-reward source transitions, and learns a simple transformation from their differences. It then applies this transformation to low-reward target transitions to generate additional target-domain data (Figure~\ref{fig:motivation}). Here, reward improvement refers to a transformation between low- and high-reward transition distributions. In this way, RIFT uses immediate reward to guide augmentation while estimating a long-horizon value function through the downstream IQL learner.

Our contributions are: (1) we introduce RIFT, which transfers the change from low- to high-reward behavior across domains and uses it to generate additional target-domain data; (2) we bound target-anchored synthesis error, give conditions for greater accuracy than source-high transplant, and connect synthesis error to a conditional downstream return guarantee; and (3) we evaluate RIFT on cross-domain locomotion tasks with changes in gravity, morphology, and friction, using approximately one million source transitions and only \(5{,}000\) target transitions, and show improved aggregate performance across several source- and target-data settings.

\section{Related Work}

\paragraph{Cross-Domain reinforcement learning.}
Reinforcement-learning policies often need to adapt across changes in embodiment, observations, action spaces, or dynamics. Prior work studies these shifts through domain randomization, system identification, meta-RL, imitation learning, and domain adaptation \citep{gamrian2019transfer,hejna2020hierarchically,eysenbach2021offdynamics,fickinger2022crossdomain}, as well as learned representations, data selection, and reward modification \citep{niu2022when,xu2023policy,guo2024darail}. In fully offline settings, DARA, BOSA, and SRPO modify rewards or constrain learning to supported regions \citep{liu2022dara,liu2024bosa,xue2023srpo}; IGDF, OTDF, and ROOT select or reweight source transitions \citep{wen2024igdf,lyu2025otdf,chen2026root}; and STC, DmC, and MOBODY correct or generate transitions \citep{yan2026stc,van2025dmc,guo2026mobody}. DROCO addresses training- and test-time dynamics robustness \citep{qiao2026droco}, while REAG and V2A align return or value information across domains \citep{wang2026reag,qiao2026v2a}. RIFT builds on this work on cross-domain transfer and data augmentation, but transfers low-to-high reward displacements learned in the source domain and applies them to target anchors to generate additional transitions.

\paragraph{Representations under domain shift.}
Domain adaptation seeks representations that preserve useful structure across domains~\citep{zhao2019invariant}. Conditional-invariance methods align distributions within labels or outcomes \citep{li2018conditional,tachetdescombes2020conditional}, including regression targets in COD~\citep{yang2024cod}. This provides a way to align task-relevant structure rather than only the overall feature distribution. Further, latent-space regularization and structured priors can also help organize learned representations \citep{higgins2017beta,shi2025gpvae}. Our method builds on this literature by using reward classes to encourage shared low- and high-reward structure across domains, supporting transfer of the improvement field.

\paragraph{Reward-structured learning and data augmentation.}
Reward and preference information can guide both policy learning and data generation. RLHF uses preferences to learn reward models or directly optimize policies \citep{ouyang2022training,gao2023scaling,rafailov2023direct}, while LENS~\citep{tao2025lens} augments limited preference data in a learned VAE latent space. In offline RL, MetaDiffuser, GTA, and GODA use return information to guide generation \citep{ni2023metadiffuser,lee2024gta,huang2025goda}, while DiffStitch connects low- and high-reward trajectory segments \citep{li2024diffstitch}. These methods use reward-related structure to construct additional data for policy learning. RIFT uses this structure for cross-domain augmentation, learning low-to-high reward transformations from source data and applying them to scarce target data. 

\section{Preliminaries}
We model each environment as a Markov decision process (MDP) $\mathcal M=(\mathcal S,\mathcal A,P,r,\gamma,\rho)$, with state space $\mathcal S$, action space $\mathcal A$, transition kernel $P$, expected one-step reward $r$, discount factor $\gamma\in[0,1)$, and initial-state distribution $\rho$. A policy $\pi$ maximizes the expected discounted return
\[
J_{\mathcal M}(\pi)
=
\mathbb E_{\rho,P,\pi}
\left[\sum_{t=0}^{T-1}\gamma^t R_t\right],
\]
where $T$ is the episode length, including the final transition, or $\infty$ for continuing tasks.

We consider source and target contexts $A$ and $B$ that share $\mathcal S$, $\mathcal A$, $\gamma$, $\rho$, and the task objective, with known termination functions. Their dynamics $P_A,P_B$ and expected rewards $r_A,r_B$ may differ: even a shared rule for scoring transitions can yield different rewards at the same $(s,a)$ when dynamics change. Given fixed offline datasets $\mathcal D_A$ and $\mathcal D_B$, with $|\mathcal D_B|\ll|\mathcal D_A|$, our goal is to learn a target policy using both datasets without further environment interaction.

\section{Methodology}
\begin{figure}[t]
    \centering
    \resizebox{\linewidth}{!}{


\begin{tikzpicture}[
    font=\small,
    >={Latex[length=2.3mm]},
    flow/.style={-{Latex}, thick},
    stage/.style={
        draw,
        rounded corners=4pt,
        thick,
        minimum width=3.8cm,
        minimum height=4.2cm,
        inner sep=6pt
    },
    smallbox/.style={
        draw,
        rounded corners=2pt,
        thick,
        minimum width=1.2cm,
        minimum height=.65cm,
        align=center
    },
    title/.style={font=\bfseries},
    tiny/.style={font=\scriptsize},
    srcLow/.style={circle, draw=blue!70!black, fill=blue!25, inner sep=1.8pt},
    srcHigh/.style={circle, draw=blue!70!black, fill=blue!75!black, inner sep=1.8pt},
    tgtLow/.style={circle, draw=orange!80!black, fill=orange!35, inner sep=1.8pt},
    tgtHigh/.style={circle, draw=orange!80!black, fill=orange!85!black, inner sep=1.8pt},
    shiftarrow/.style={-{Latex[length=1.6mm]}, draw=gray!55, line width=0.8pt},
]







\node[stage] (s1) at (2.5,0) {};

\node[title] at ($(s1.north)+(0,-0.35)$) {1. Encode};

\draw[gray!70, thick, -{Latex[length=2mm]}]
    ($(s1.center)+(-1.25,0.95)$) -- ($(s1.center)+(1.25,0.95)$);

\node[tiny] at ($(s1.center)+(-0.95,1.18)$) {low reward};
\node[tiny] at ($(s1.center)+( 0.95,1.18)$) {high reward};


\node[srcLow]  at ($(s1.center)+(-1.05, 0.32)$) {};
\node[srcLow]  at ($(s1.center)+(-0.88, 0.12)$) {};
\node[srcLow]  at ($(s1.center)+(-0.72, 0.28)$) {};
\node[srcLow]  at ($(s1.center)+(-0.58, 0.02)$) {};
\node[srcLow]  at ($(s1.center)+(-0.92,-0.08)$) {};
\node[srcLow]  at ($(s1.center)+(-0.70,-0.18)$) {};

\node[srcHigh] at ($(s1.center)+( 0.42, 0.26)$) {};
\node[srcHigh] at ($(s1.center)+( 0.60, 0.42)$) {};
\node[srcHigh] at ($(s1.center)+( 0.78, 0.18)$) {};
\node[srcHigh] at ($(s1.center)+( 0.92, 0.34)$) {};
\node[srcHigh] at ($(s1.center)+( 1.08, 0.08)$) {};
\node[srcHigh] at ($(s1.center)+( 0.72,-0.08)$) {};

\node[tgtLow]  at ($(s1.center)+(-0.98, 0.02)$) {};
\node[tgtLow]  at ($(s1.center)+(-0.50,-0.28)$) {};
\node[tgtLow]  at ($(s1.center)+(-0.88,-0.38)$) {};

\node[tgtHigh] at ($(s1.center)+( 0.68,-0.18)$) {};
\node[tgtHigh] at ($(s1.center)+( 1.00 ,0.06)$) {};
\node[tgtHigh] at ($(s1.center)+( 0.90,-0.48)$) {};


\node[tiny] at ($(s1.south)+(0,0.35)$)
    {\textcolor{blue!70!black}{$\bullet$}\ $\mathcal D_A$
     \hspace{1.0em}
     \textcolor{orange!80!black}{$\bullet$}\ $\mathcal D_B$};


\node[stage, right=1.3cm of s1] (s2) {};

\node[title] at ($(s2.north)+(0,-0.35)$) {2. Learn shift};

\node[tiny] at ($(s2.center)+(-0.95,0.78)$) {low};
\node[tiny] at ($(s2.center)+( 0.95,0.78)$) {high};

\node[srcLow]  at ($(s2.center)+(-1.00, 0.32)$) {};
\node[srcLow]  at ($(s2.center)+(-0.82, 0.10)$) {};
\node[srcLow]  at ($(s2.center)+(-0.66, 0.26)$) {};
\node[srcLow]  at ($(s2.center)+(-0.88,-0.18)$) {};
\node[srcLow]  at ($(s2.center)+(-0.62,-0.06)$) {};

\node[srcHigh] at ($(s2.center)+( 0.55, 0.28)$) {};
\node[srcHigh] at ($(s2.center)+( 0.78, 0.10)$) {};
\node[srcHigh] at ($(s2.center)+( 0.98, 0.30)$) {};
\node[srcHigh] at ($(s2.center)+( 0.70,-0.12)$) {};
\node[srcHigh] at ($(s2.center)+( 0.92,-0.02)$) {};

\draw[shiftarrow] ($(s2.center)+(-0.90,0.30)$) -- ($(s2.center)+(0.42,0.28)$);
\draw[shiftarrow] ($(s2.center)+(-0.74,0.08)$) -- ($(s2.center)+(0.62,0.10)$);
\draw[shiftarrow] ($(s2.center)+(-0.80,-0.16)$) -- ($(s2.center)+(0.56,-0.12)$);


\node at ($(s2.center)+(0,-0.95)$)
    {$\hat{\delta}(u)=Wu+b$};

\draw[flow] (s1.east) -- (s2.west);

\node[tiny] at ($(s2.south)+(0,0.35)$)
    {\textcolor{blue!70!black}{$\bullet$}\ $\mathcal D_A$};


\node[stage, right=1.3cm of s2] (s3) {};

\node[title] at ($(s3.north)+(0,-0.35)$)
    {3. Transfer};

\node[tiny, text=gray!70] at ($(s3.center)+(0,0.95)$)
    {$\hat{\delta}(u)=Wu+b$};

\node[tgtLow] at ($(s3.center)+(-0.98, 0.30)$) {};
\node[tgtLow] at ($(s3.center)+(-0.82,-0.02)$) {};
w\node[tgtLow] at ($(s3.center)+(-0.96,-0.34)$) {};

\draw[orange!80!black, thick, fill=white]
    ($(s3.center)+(0.58, 0.27)$) circle (2.3pt);

\draw[orange!80!black, thick, fill=white]
    ($(s3.center)+(0.86, 0.10)$) circle (2.3pt);

\draw[orange!80!black, thick, fill=white]
    ($(s3.center)+(0.72,-0.20)$) circle (2.3pt);

\draw[shiftarrow]
    ($(s3.center)+(-0.90, 0.30)$)
    -- ($(s3.center)+(0.46, 0.27)$);

\draw[shiftarrow]
    ($(s3.center)+(-0.74,-0.02)$)
    -- ($(s3.center)+(0.72, 0.08)$);

\draw[shiftarrow]
    ($(s3.center)+(-0.88,-0.34)$)
    -- ($(s3.center)+(0.60,-0.18)$);
    

\draw[flow] (s2.east) -- (s3.west);

\node[tiny] at ($(s3.south)+(0,0.35)$)
    {\textcolor{orange!80!black}{$\bullet$}\ $\mathcal D_B^{\mathrm{low}}$
     \hspace{0.9em}
     \textcolor{orange!80!black}{$\circ$}\ $\mathcal D'_B$};


\node[stage, right=1.3cm of s3] (s4) {};

\node[title] at ($(s4.north)+(0,-0.35)$)
    {4. Train};


\node[smallbox, minimum width=1.05cm] (da)
    at ($(s4.center)+(-1.15,0.65)$)
    {$\mathcal D_A$};

\node[smallbox, minimum width=1.05cm] (db)
    at ($(s4.center)+(0,0.65)$)
    {$\mathcal D_B$};

\node[smallbox, minimum width=1.05cm] (dp)
    at ($(s4.center)+(1.15,0.65)$)
    {$\mathcal D'_B$};


\coordinate (joinL) at ($(s4.center)+(-1.15,-0.05)$);
\coordinate (joinM) at ($(s4.center)+(0,-0.05)$);
\coordinate (joinR) at ($(s4.center)+(1.15,-0.05)$);
\coordinate (mergeTrain) at ($(s4.center)+(0,-0.35)$);

\draw[thick] (da.south) -- (joinL);
\draw[thick] (db.south) -- (joinM);
\draw[thick] (dp.south) -- (joinR);

\draw[thick] (joinL) -- (joinR);

\node[smallbox, minimum width=1.6cm, minimum height=.75cm] (iql)
    at ($(s4.center)+(0,-1.05)$)
    {IQL};

\draw[flow] (joinM) -- (iql.north);

\draw[flow] (s3.east) -- (s4.west);

\end{tikzpicture}

}
    \caption{Overview of RIFT. The method encodes source and target data, learns a reward-improvement field from source transitions, transfers it to target anchors, and augments downstream IQL training.}
    \label{fig:method}
\end{figure}
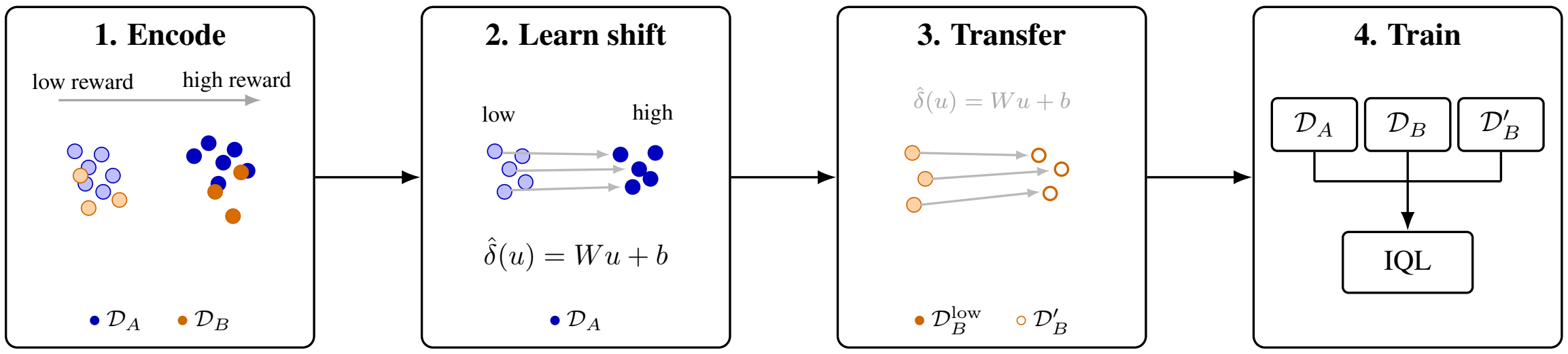

Reward Improvement Field Transfer (RIFT) learns a low-to-high reward transformation from abundant source data in context $A$ and applies it to scarce target data in context $B$ to generate additional transitions (Figure~\ref{fig:method}). This design addresses target-data scarcity: $\mathcal D_B$ may be too small to fit a reliable dynamics model or a flexible transition generator, motivating a simple source-learned transformation applied to target anchors.

\subsection{Algorithm}
\label{sec:alg}

First, we learn a shared latent representation from $\mathcal D_{\mathrm{mix}}=\mathcal D_A\cup\mathcal D_B$. Within each context, we label transitions in the bottom and top reward quantiles:
\[
y(x)=
\begin{cases}
0, & r_c\leq Q_c(q),\\
1, & r_c\geq Q_c(1-q),
\end{cases}
\qquad c\in\{A,B\},
\]
where $Q_c$ is the reward quantile function and $q\in(0,1/2)$ is fixed. Smaller $q$ increases or preserves the reward separation between classes but leaves fewer samples for field fitting and target anchoring.

We then train a shared variational autoencoder (VAE) \citep{kingma2014autoencoding} on transitions $x=(s_t,a_t,s_{t+1})$. Alongside the standard normal prior $\mathcal N(0,I)$, we add class-conditioned regularization on one latent coordinate, $z_0,$ to encourage a shared low-to-high reward direction: 
\[
p(z_0\mid y=0)=\mathcal N(-c_q,s_q^2),
\qquad
p(z_0\mid y=1)=\mathcal N(c_q,s_q^2).
\]
These means and variance match those of the lower and upper $q$-tails of a standard normal and are therefore determined by $q$. After training, the encoder posterior mean $z=e_\vartheta(x)$ represents each transition, and the decoder $g_\vartheta$ maps latent points back to transitions.

Next, we learn a low-to-high reward transformation from source context $A$. We append the observed reward to each latent point, giving $u=(z,r)$, and normalize source points to target coordinates:
\[
\tilde u
=
\bm\mu_B+
\bm\sigma_B\cdot
\frac{u-\bm\mu_A}{\bm\sigma_A},
\]
where $\bm\mu_c$ and $\bm\sigma_c$ are the coordinate-wise means and standard deviations over all samples in context $c$. Mini-batch optimal transport~\citep{villani2009optimal} then pairs these normalized low- and high-reward source points using squared Euclidean distance between their latent components. For each matched low/high pair $(\tilde u_i,\tilde v_i)$, we compute the joint latent--reward displacement $\delta_i=\tilde v_i-\tilde u_i$. We fit an affine field $\widehat\delta(u)=\widehat W u+\widehat b$ to these displacements by ordinary least squares. For each low-reward target anchor $u^B$,
we generate 
\[
\widetilde u^B=u^B+\widehat\delta(u^B).
\]
We decode its latent component into a transition and use its reward component as the synthetic reward. The known target termination function assigns the terminal flag from the decoded
next state, and samples with terminal initial states are discarded.

Finally, we train the downstream policy with implicit Q-learning (IQL)~\citep{kostrikov2022offline} and evaluate it in context $B$. While immediate reward provides the local signal for augmentation, downstream IQL learns the long-horizon values through Bellman backups. We standardize observations using means and standard deviations. Further, each minibatch draws a fraction $1-\lambda$ of its transitions uniformly from $\mathcal D_A$ and the remaining fraction $\lambda$ uniformly from $\mathcal D_B\cup\mathcal D_B'$, where $\mathcal D_B'$ contains RIFT's synthetic target transitions. The weight $\lambda$ balances source coverage against source reward and dynamics mismatch. The realized synthetic fraction within the target pool is $\theta=|\mathcal D_B'|/(|\mathcal D_B|+|\mathcal D_B'|)$.

\subsection{Theoretical Analysis}
\label{sec:theory}
We formalize when a source improvement field can transfer across domains, bound synthesis error, and compare target anchoring with directly decoding high-reward source points. Proofs and the conditional downstream guarantee appear in Appendix~\ref{app:add_proofs}. All appendices are provided in the supplementary material.

\paragraph{Setting.}\label{sec:setting}
Let $\eta_c^y$ denote the transition--reward distribution in context
$c$ and immediate-reward class $y$. Define
\[
E(x,r)=(e_\vartheta(x),r),\qquad
D(z,r)=(g_\vartheta(z),r),\qquad
\mu_c^y=E_\#\eta_c^y,
\]
where $\#$ denotes the pushforward of a distribution. Write $\bar\mu_A^y=S_\#\mu_A^y$ for source points normalized to target coordinates as in Section~\ref{sec:alg}. We fix the representation, normalization, and fitted field. All norms and Wasserstein distances use Euclidean coordinates, and all distributions have finite second moments.

Let $\pi_A\in\Pi(\bar\mu_A^0,\bar\mu_A^1)$ be a population coupling that minimizes squared distance between the latent components. Its population affine fit is
\[
\delta_{\rm aff}(u)=Wu+b
\in\arg\min_{\delta\text{ affine}}
\mathbb E_{(X,Y)\sim\pi_A}
\|Y-X-\delta(X)\|^2.
\]
Choose one minimizer if necessary. The fitted generator and its pre-filtering synthetic distribution are
\[
F=\mathrm{id}+\widehat\delta,
\qquad
\nu=(D\circ F)_\#\mu_B^0.
\]

\begin{riftassumption}[Shared class geometry]\label{ass:transfer}
There is a common invertible affine map $h(u)=Hu+t$ such that
\[
W_2(\mu_B^y,h_\#\bar\mu_A^y)\le\rho_y,
\qquad y=0,1.
\]
\end{riftassumption}

The map $h$ describes the remaining cross-context geometry after normalization. The reward-conditioned VAE regularization is designed to  encourage this correspondence.

\paragraph{Synthesis error.}
Define the field mismatch
$\Gamma_h(u)=\delta_{\rm aff}(u)
-H\delta_{\rm aff}(h^{-1}(u))$ and the errors
\[
\begin{aligned}
\varepsilon_{\rm aff}
&=\bigl(\mathbb E_{\pi_A}
\|Y-X-\delta_{\rm aff}(X)\|^2\bigr)^{1/2},
&
\varepsilon_{\rm est}
&=\|\widehat\delta-\delta_{\rm aff}\|_{L^2(\mu_B^0)},\\
\varepsilon_{\rm conj}
&=\|\Gamma_h\|_{L^2(\mu_B^0)},
&
\varepsilon_{\rm dec}
&=\bigl(\mathbb E_{(x,r)\sim\eta_B^1}
\|g_\vartheta(e_\vartheta(x))-x\|^2\bigr)^{1/2}.
\end{aligned}
\]
These measure source affine approximation, field estimation on target
anchors, cross-context field mismatch, and target-high reconstruction,
respectively. Let $L_h=\|H(I+W)H^{-1}\|$ and
$L_D=\max(1,L_g)$, where the decoder is $L_g$-Lipschitz.
The population transfer-error budget is then
\[
\varepsilon_{\rm tr}
=
\varepsilon_{\rm conj}
+\|H\|\varepsilon_{\rm aff}
+L_h\rho_0+\rho_1.
\]

\begin{rifttheorem}[Accuracy of transferred synthesis]
\label{thm:synthesis}
Under Assumption~\ref{ass:transfer} and the setting above,
\begin{equation}\label{eq:synthesis-bound}
W_2(\nu,\eta_B^1)
\le
\varepsilon_{\rm syn}
:=
L_D(\varepsilon_{\rm est}+\varepsilon_{\rm tr})
+\varepsilon_{\rm dec}.
\end{equation}
Thus, generated and genuine target-high points admit a coupling with root-mean-square distance at most $\varepsilon_{\rm syn}$.
\end{rifttheorem}

For a residual translation $h(u)=u+t$,
$\varepsilon_{\rm conj}=\|Wt\|$. A large common shift can therefore produce little field mismatch when the improvement field is nearly constant.

\begin{riftproposition}[Comparison with source-high transplant]\label{prop:transplant}
Suppose the encoder is $L_e$-Lipschitz, put $L_E=\max(1,L_e)$, and let $\varepsilon_{\rm cyc}=(\mathbb E_{(z,r)\sim\bar\mu_A^1}\|e_\vartheta(g_\vartheta(z))-z\|^2)^{1/2}$. With finite second moments,
\begin{equation}\label{eq:transplant-bound}
W_2(D_\#\bar\mu_A^1,\eta_B^1)
\ge\frac{W_2(\bar\mu_A^1,\mu_B^1)-\varepsilon_{\rm cyc}}{L_E}.
\end{equation}
If this lower bound exceeds $\varepsilon_{\rm syn}$ under Theorem~\ref{thm:synthesis}'s assumptions, RIFT is strictly closer to $\eta_B^1$ than transplant is. Under Assumption~\ref{ass:transfer} with $h(u)=u+t$, transplant's lower bound is at least $(\|t\|-\rho_1-\varepsilon_{\rm cyc})/L_E$, whereas RIFT's field-incompatibility contribution is $L_D\|Wt\|$.
\end{riftproposition}

\paragraph{From source fitting to synthesis accuracy.}
Under the regularity, population-coupling, and sample-size conditions in Appendix~\ref{app:estimation}, the fitted field satisfies
\[
\varepsilon_{\rm est}
\le
\overline{\varepsilon}_{\rm est}(J)
:=
\kappa_M
\left[B_k+C_2\sqrt{\frac{\log(2/\eta)}{J}}\right]
\]
with probability at least $1-\eta$, where $J$ is the number of independent source batches, $B_k$ bounds matching bias for batches of $k$ points per class, and $\kappa_M$ converts source prediction error to error on target anchors. The constant $C_2$ depends on source boundedness
and moment conditioning.

Substituting into Theorem~\ref{thm:synthesis} gives
\begin{equation}\label{eq:finite-source-synthesis}
W_2(\nu,\eta_B^1)
\le
L_D\bigl[\overline{\varepsilon}_{\rm est}(J)
+\varepsilon_{\rm tr}\bigr]+\varepsilon_{\rm dec}
\end{equation} 
with the same probability. If this upper bound is below the transplant lower bound in \eqref{eq:transplant-bound}, the strict comparison therefore holds for the estimated field.

Increasing $J$ reduces sampling error, but not matching bias, transfer error, or decoding error. If $L_D(\kappa_M B_k+\varepsilon_{\rm tr})+\varepsilon_{\rm dec}$ is below the transplant lower bound, sufficiently many independent source batches establish the comparison for fixed $k$ and $\eta$. This motivates learning an improvement rule from abundant source data, while recognizing that more source data alone cannot resolve field mismatch.

\paragraph{Sampling interpretation.}
Theorem~\ref{thm:synthesis} concerns the pre-filtering distribution $\nu$, not a finite generated dataset. When $\varepsilon_{\rm syn}=0$, independent target-low anchors
passed through the fixed generator produce independent target-high samples. Finite datasets still have sampling error, which we analyze in Appendix~\ref{app:sampling}.

\paragraph{Conditional downstream consequence.}
Under target and surrogate regularity, occupancy domination, and a surrogate learning-error allowance $\varepsilon_{\rm learn}$,
Theorem~\ref{thm:policy} in Appendix~\ref{app:policy} gives
\[
J_B(\widehat\pi)
\ge J_B(\pi_c)-\varepsilon_{\rm learn}
-\mathcal R_{\lambda,\theta}(\widehat\pi)
-\mathcal R_{\lambda,\theta}(\pi_c).
\]
The budget $\mathcal R$ includes synthesis error, source reward and dynamics mismatch, filtering, and surrogate fitting, with terminal-flag errors included for episodic tasks. This connects synthesis accuracy to return, motivating our overall approach.

\section{Experiments}
We evaluate RIFT on the ODRL cross-domain offline RL benchmark~\citep{lyu2024odrl} and use ablations to examine its components and hyperparameters. Hyperparameters and additional
results appear in Appendices~\ref{app:hyperparams}
and~\ref{app:results}.

\subsection{Setup}
\paragraph{Tasks and evaluation protocol.}
We evaluate HalfCheetah, Hopper, Walker2d, and Ant under ODRL's gravity, morphology, and friction shifts, following the benchmark
settings used by OTDF~\citep{lyu2025otdf} and STC~\citep{yan2026stc}. Target gravity and friction coefficients are halved in their respective settings; the medium morphology shift modifies one body component per task. We use the full source dataset $\mathcal D_A$, containing approximately one million transitions, and $5{,}000$ target transitions in $\mathcal D_B$, held fixed across policy seeds. Dataset qualities follow D4RL \citep{fu2020d4rl}: medium (\texttt{med}), medium-replay (\texttt{med-r}), medium-expert (\texttt{med-e}), and expert.
Further dataset details appear in Appendix~\ref{app:datasets}.

\paragraph{Baselines.}
We compare against pooled source--target IQL
\citep{kostrikov2022offline} and six cross-domain methods. DARA adjusts source rewards using domain classifiers \citep{liu2022dara}, BOSA constrains policy and value learning to supported regions \citep{liu2024bosa}, and SRPO shapes rewards using stationary-state information \citep{xue2023srpo}. IGDF uses contrastive scores to select source transitions \citep{wen2024igdf}, while OTDF scores and filters them using optimal transport \citep{lyu2025otdf}. STC corrects source transitions toward the target domain and filters unreliable corrections \citep{yan2026stc}. Baseline results in the source- and target-quality comparisons are taken from STC under the same ODRL protocol, using one million gradient steps and five seeds. Implementation details appear in Appendix~\ref{app:hyperparams}.

\subsection{Main Results}\label{sec:main-results}

\paragraph{Performance across source qualities.}
Table~\ref{tab:source-quality} varies source-data quality while fixing the target dataset to $5{,}000$ medium-quality transitions. RIFT achieves the highest mean in $16$ of $24$ gravity and morphology settings and the highest aggregate normalized score of $1318.7$, a $26.2\%$ gain over STC ($1045.3$), the strongest baseline in aggregate. Gains are largest on HalfCheetah and Hopper, particularly under morphology shifts, and less consistent on Walker2d and Ant. OTDF and STC generally outperform earlier adaptation methods, highlighting the value of addressing transition mismatch. RIFT further improves aggregate performance by learning low-to-high reward transformations from source data.

\paragraph{Performance across target qualities.}
We also fix source quality to medium and varies target quality (Appendix~\ref{app:target-quality}). RIFT achieves the highest mean in $13$ of $20$ settings and the highest aggregate score of $940.9$, a $14.6\%$ improvement over STC ($820.9$). RIFT's advantage is generally larger with medium-quality target data and decreases as target quality improves. This is consistent with its motivation: augmentation offers more benefit when useful target behavior is not already well represented in the dataset.

\begin{table}[t]
\centering
\caption{Performance under gravity and morphology shifts when the target dataset contains medium-quality transitions and source-data quality varies. Values report the mean $\pm$ standard deviation over five seeds, with the highest mean bolded.}
\label{tab:source-quality}
\setlength{\tabcolsep}{3pt}
\resizebox{\textwidth}{!}{%
\begin{tabular}{lll*{8}{c}}
\toprule
Type & Source & Target
& IQL & DARA & BOSA & SRPO & IGDF & OTDF & STC
& \textbf{RIFT (ours)} \\
\midrule
Gravity & half-med & med
& \score{39.6}{3.3} & \score{41.2}{3.9} & \score{38.9}{4.0}
& \score{36.9}{4.5} & \score{36.6}{5.5} & \score{40.7}{7.7}
& \score{42.4}{5.3} & \bestscore{47.3}{6.6} \\
& half-med-r & med
& \score{20.1}{5.0} & \score{17.6}{6.2} & \score{20.0}{4.9}
& \score{17.5}{5.2} & \score{14.4}{2.2} & \score{21.5}{6.5}
& \score{26.7}{2.2} & \bestscore{31.4}{5.3} \\
& half-med-e & med
& \score{38.6}{6.0} & \score{37.8}{3.3} & \score{41.8}{5.1}
& \score{42.5}{2.3} & \score{37.7}{7.3} & \score{39.5}{3.5}
& \score{39.2}{4.2} & \bestscore{49.6}{5.3} \\
& hopp-med & med
& \score{11.2}{1.1} & \score{17.3}{3.8} & \score{15.2}{3.3}
& \score{12.4}{1.0} & \score{15.3}{3.5} & \score{32.4}{8.0}
& \score{43.4}{6.1} & \bestscore{60.4}{7.3} \\
& hopp-med-r & med
& \score{13.9}{2.9} & \score{10.7}{4.3} & \score{3.3}{1.9}
& \score{14.0}{2.6} & \score{15.3}{4.4} & \score{31.1}{13.4}
& \score{36.8}{17.8} & \bestscore{61.3}{4.3} \\
& hopp-med-e & med
& \score{19.1}{6.6} & \score{18.5}{12.3} & \score{15.9}{5.9}
& \score{19.7}{8.5} & \score{22.3}{5.4} & \score{26.4}{10.1}
& \score{45.3}{7.5} & \bestscore{65.2}{6.5} \\
& walk-med & med
& \score{28.1}{12.9} & \score{28.4}{13.7} & \score{38.0}{11.2}
& \score{21.4}{7.0} & \score{22.1}{8.4} & \score{36.6}{2.3}
& \score{41.6}{4.0} & \bestscore{46.5}{5.5} \\
& walk-med-r & med
& \score{14.6}{2.5} & \score{14.1}{6.1} & \score{7.6}{5.8}
& \score{17.9}{3.8} & \score{11.6}{4.6} & \score{32.7}{7.0}
& \score{29.0}{1.9} & \bestscore{33.0}{8.8} \\
& walk-med-e & med
& \score{39.9}{13.1} & \score{41.6}{13.0} & \score{32.3}{7.2}
& \bestscore{46.4}{3.5} & \score{33.8}{3.1} & \score{30.2}{9.8}
& \score{34.9}{9.1} & \score{40.3}{7.6} \\
& ant-med & med
& \score{10.2}{1.8} & \score{9.4}{0.9} & \score{12.4}{2.0}
& \score{11.7}{1.0} & \score{11.3}{1.3} & \bestscore{45.1}{12.4}
& \score{42.6}{8.4} & \score{40.8}{10.3} \\
& ant-med-r & med
& \score{18.9}{2.6} & \score{21.7}{2.1} & \score{13.9}{1.5}
& \score{18.7}{1.7} & \score{19.6}{1.0} & \score{29.6}{10.7}
& \score{40.9}{5.6} & \bestscore{46.7}{3.5} \\
& ant-med-e & med
& \score{9.8}{2.4} & \score{8.1}{1.8} & \score{8.1}{3.0}
& \score{8.4}{2.1} & \score{8.9}{1.5} & \score{18.6}{11.9}
& \bestscore{39.2}{9.2} & \score{38.1}{7.0} \\
\midrule
Morph & half-med & med
& \score{24.5}{2.4} & \score{21.0}{3.9} & \score{24.2}{5.6}
& \score{18.1}{1.8} & \score{23.7}{3.4} & \score{21.1}{7.6}
& \score{19.5}{2.2} & \bestscore{57.9}{4.4} \\
& half-med-r & med
& \score{11.0}{1.2} & \score{9.5}{2.3} & \score{4.7}{2.9}
& \score{8.9}{1.2} & \score{9.2}{0.6} & \score{6.5}{1.4}
& \score{13.0}{5.3} & \bestscore{49.2}{4.6} \\
& half-med-e & med
& \score{21.1}{2.8} & \score{19.2}{2.2} & \score{23.2}{3.9}
& \score{21.1}{1.9} & \score{18.6}{1.3} & \score{20.8}{2.5}
& \score{16.8}{8.8} & \bestscore{60.5}{3.1} \\
& hopp-med & med
& \score{15.9}{6.8} & \score{17.8}{10.1} & \score{12.8}{0.1}
& \score{21.7}{7.7} & \score{25.3}{9.7} & \score{16.4}{7.1}
& \score{43.1}{23.9} & \bestscore{69.4}{13.2} \\
& hopp-med-r & med
& \score{12.9}{0.3} & \score{12.8}{0.1} & \score{2.0}{1.2}
& \score{12.4}{0.7} & \score{12.5}{1.7} & \score{13.3}{0.1}
& \score{22.9}{6.1} & \bestscore{61.8}{10.6} \\
& hopp-med-e & med
& \score{14.9}{3.1} & \score{11.1}{5.6} & \score{14.4}{1.8}
& \score{16.6}{1.9} & \score{18.3}{7.5} & \score{25.4}{9.4}
& \score{53.4}{20.8} & \bestscore{65.7}{10.4} \\
& walk-med & med
& \score{31.5}{8.6} & \score{35.0}{10.8} & \score{26.7}{6.6}
& \score{38.6}{5.1} & \score{38.5}{8.4} & \score{42.5}{3.1}
& \bestscore{56.7}{8.1} & \score{52.8}{8.2} \\
& walk-med-r & med
& \score{41.5}{3.0} & \score{38.5}{7.9} & \score{15.3}{7.6}
& \score{36.0}{4.4} & \score{24.2}{8.6} & \score{17.9}{13.4}
& \bestscore{63.1}{8.0} & \score{46.9}{6.7} \\
& walk-med-e & med
& \score{32.8}{4.3} & \score{41.4}{10.9} & \score{45.1}{13.7}
& \score{39.8}{14.3} & \score{37.9}{4.2} & \score{55.3}{2.2}
& \bestscore{62.1}{8.1} & \score{61.3}{7.0} \\
& ant-med & med
& \score{71.4}{2.4} & \score{71.5}{6.8} & \score{54.8}{13.2}
& \score{72.8}{2.2} & \score{71.8}{2.7} & \score{75.1}{2.3}
& \bestscore{77.2}{2.8} & \score{76.2}{2.4} \\
& ant-med-r & med
& \score{65.9}{5.5} & \score{62.3}{8.2} & \score{15.2}{2.3}
& \score{59.3}{4.0} & \score{65.0}{5.3} & \score{63.1}{5.9}
& \score{76.2}{2.6} & \bestscore{78.1}{0.4} \\
& ant-med-e & med
& \score{70.2}{7.3} & \score{64.3}{5.8} & \score{64.0}{6.0}
& \score{68.5}{4.4} & \score{66.8}{11.0} & \score{76.4}{1.9}
& \bestscore{79.3}{0.2} & \score{78.3}{0.3} \\
\midrule
\multicolumn{3}{l}{Total (displayed means)}
& 677.6 & 670.8 & 549.8 & 681.3 & 660.7 & 818.2 & 1045.3 & \textbf{1318.7} \\
\bottomrule
\end{tabular}%
}
\end{table}


\subsection{Ablations}
We examine augmentation, representation design, and sensitivity to reward conditioning and target-data size across eight medium-to-medium gravity and morphology settings. All experiments follow the main protocol except for the component varied. Our reproduced baseline, IQL*, uses the same selection protocol as RIFT: for each task, we report the choice of $\lambda\in\{0.5,1\}$ with the higher mean target return. This distinguishes it from the published IQL results used in our main comparisons.

\paragraph{Augmentation and source-field controls.}
Table~\ref{tab:ablation-augmenter} compares RIFT with
Mixup~\citep{zhang2018mixup} and SMOTE~\citep{chawla2002smote},
using either target data alone or pooled source and target data. Each uses a 1,000-transition budget. RIFT achieves the highest aggregate score
of $451.3$, compared with $394.5$ for the strongest interpolation baseline, although results vary by task. We also compare against a target-only field using the same VAE and anchors, and resampling real high-reward target transitions. These score $423.8$ and $384.5$, respectively, supporting the
benefit of learning the field from source data
(Appendix~\ref{app:mechanism-controls}).

\begin{table}[t]
\centering
\small
\caption{Comparison of augmentation strategies across eight medium-to-medium gravity and morphology settings. Mixup and SMOTE generate 1,000 synthetic transitions using either pooled source and target data ($A\cup B$) or target data alone ($B$). RIFT generates up to 1,000 synthetic transitions. Values report the mean $\pm$ standard deviation over five seeds, with the highest mean bolded.}
\label{tab:ablation-augmenter}
\setlength{\tabcolsep}{4pt}
\begin{adjustbox}{max width=\textwidth}
\begin{tabular}{llcccccc}
\toprule
& & & \multicolumn{2}{c}{Mixup}
& \multicolumn{2}{c}{SMOTE} & \\
\cmidrule(lr){4-5}\cmidrule(lr){6-7}
Shift & Task & IQL*
& $A\cup B$ & $B$ & $A\cup B$ & $B$ & RIFT \\
\midrule
Gravity & Ant
& \score{29.8}{4.9}
& \score{29.5}{1.9}
& \score{26.1}{6.9}
& \score{28.2}{7.5}
& \score{34.4}{7.2}
& \bestscore{40.8}{10.3} \\
& HalfCheetah
& \score{42.0}{2.3}
& \score{42.9}{2.1}
& \score{43.3}{2.0}
& \score{45.2}{4.7}
& \score{45.4}{4.9}
& \bestscore{47.3}{6.6} \\
& Hopper
& \score{52.7}{12.5}
& \score{49.1}{12.1}
& \score{44.1}{12.8}
& \bestscore{61.0}{6.6}
& \score{60.8}{8.6}
& \score{60.4}{7.3} \\
& Walker2d
& \score{30.6}{4.3}
& \score{43.9}{6.1}
& \score{36.5}{13.6}
& \score{45.1}{6.6}
& \score{40.5}{5.5}
& \bestscore{46.5}{5.5} \\
\midrule
Morphology & Ant
& \score{76.9}{0.3}
& \score{74.1}{2.8}
& \score{76.2}{0.7}
& \bestscore{77.3}{0.5}
& \bestscore{77.3}{0.3}
& \score{76.2}{2.4} \\
& HalfCheetah
& \score{22.8}{2.4}
& \score{18.8}{1.5}
& \score{17.3}{1.8}
& \score{18.8}{2.5}
& \score{20.2}{4.8}
& \bestscore{57.9}{4.4} \\
& Hopper
& \score{55.5}{6.4}
& \score{59.9}{13.9}
& \score{46.6}{10.3}
& \score{54.3}{12.1}
& \score{60.2}{11.2}
& \bestscore{69.4}{13.2} \\
& Walker2d
& \score{49.2}{5.1}
& \score{54.1}{8.6}
& \bestscore{59.0}{9.4}
& \score{54.3}{7.8}
& \score{55.7}{10.9}
& \score{52.8}{8.2} \\
\midrule
\multicolumn{2}{l}{Total (displayed means)}
& 359.5 & 372.3 & 349.1 & 384.2 & 394.5
& \textbf{451.3} \\
\bottomrule
\end{tabular}
\end{adjustbox}
\end{table}

\paragraph{Representation design.}
Balanced source--target VAE sampling improves the aggregate score from $394.6$ to $432.7$, and adding reward regularization raises it to $451.3$ (Appendix~\ref{app:vae-design}). For the quantile, we chose $q=0.2$ as the default before running this ablation. It also achieves the highest aggregate score among the tested quantiles, although the preferred value varies by task (Appendix~\ref{app:reward-quantile}).

\paragraph{Conditioning horizon and target-data size.}
Using 5-, 10-, and 20-step-return classes gives broadly similar aggregate performance, with no consistent gain as the horizon increases (Appendix~\ref{app:reward-horizon}). This supports immediate reward in these dense-reward tasks while showing that RIFT accommodates longer-horizon signals, which may be useful for sparse or delayed rewards. Further, RIFT improves aggregate performance over IQL* with 1,000 and 2,500 target transitions, but underperforms at 500, possibly because limited target coverage weakens representation learning and target anchoring (Appendix~\ref{app:target-size}).

\subsection{Latent Space Representation}
\label{sec:latent-space}

\begin{figure}[t]
    \centering
    \includegraphics[width=0.65\textwidth]{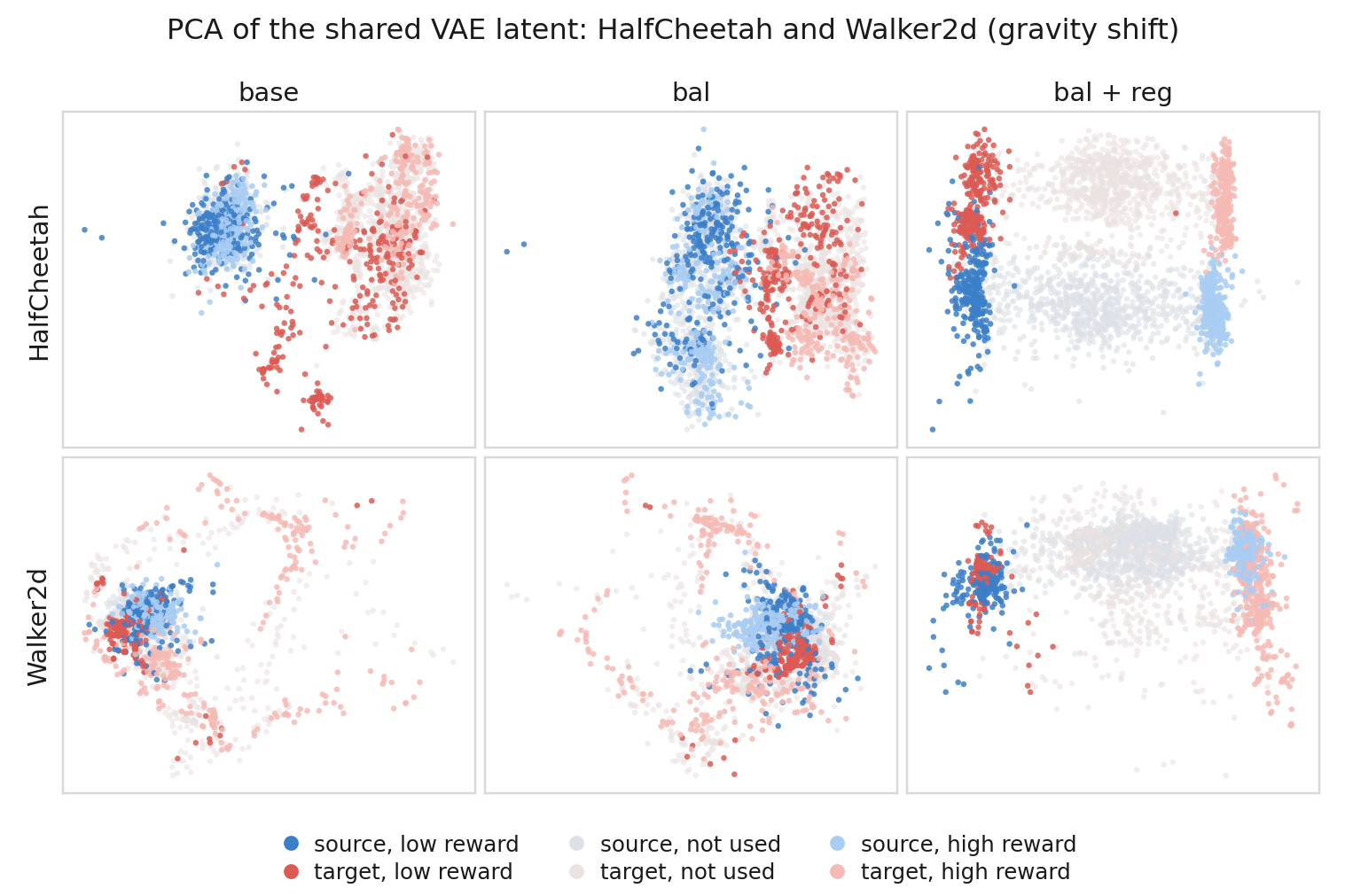}
    \caption{PCA projections of shared VAE representations under
    gravity shifts. Colors distinguish domains and reward classes.
    The regularized representation shows a similar low-to-high
    reward direction across domains.}
    \label{fig:latent-pca}
\end{figure}

Figure~\ref{fig:latent-pca} shows shared VAE representations for HalfCheetah and Walker2d under gravity shifts. Balanced sampling and reward regularization produce a similar low-to-high reward direction across domains, consistent with the shared-geometry motivation. Reward regularization also preserves reconstruction quality, explaining over $98\%$ of the variance in all settings (Appendix~\ref{app:synthetic-validity}).

We also evaluate held-out reconstruction and replay generated transitions in the target simulator. Per-dimension reconstruction MSE ranges from 0.001--0.07 times the corresponding variance, while reward errors range from 0.05--0.19 target reward-class gaps. These results support reconstruction quality and the accuracy of generated reward labels, although next-state accuracy varies by task and does not consistently track policy performance (Appendix~\ref{app:synthetic-validity}).

\section{Conclusions}

We introduced RIFT, which learns low-to-high reward transformations from abundant source data and applies them to target anchors to generate additional training transitions. Our analysis bounds synthesis error, identifies conditions favoring target anchoring over source-high transplant, and gives a conditional connection to target return. Experiments under gravity, morphology, and friction shifts show improved aggregate performance across source- and target-data qualities. Controlled comparisons further support the contribution of the source-derived field beyond high-reward target resampling or learning the field from target data alone. These findings suggest that source data can provide useful relationships between transitions even when the transitions themselves do not match target dynamics. Transferring these relationships, specifically for reward improvement, offers a way to use existing experience when collecting additional target data is difficult or expensive.

RIFT also has limitations. Reward regularization encourages the source field to remain useful in the target domain but does not guarantee compatibility or detect failure under larger dynamics shifts. Our evaluation focuses on dense-reward locomotion; sparse or delayed rewards may require other conditioning signals. Longer-horizon labels offer one option, but evaluation on these tasks and manipulation environments remains future work. Finally, the theory fixes the representation and population coupling under additional regularity and sampling assumptions;
it does not prove that end-to-end training satisfies them or that accurate synthesis alone guarantees improved return.

\subsection*{Generative AI Use}
We used generative AI tools for writing and LaTeX editing, code development, related-work search, and assistance with mathematical formalization, proof writing, and consistency checks. The core research ideas and mathematical reasoning were developed by the authors. We reviewed the AI-assisted text, code, proofs, and references and take responsibility for the final content.

\bibliography{l4dc2027-sample}

\clearpage
\appendix
\section{Theoretical Results and Proofs}\label{app:add_proofs}
\setcounter{rifttheorem}{0}\renewcommand{\therifttheorem}{\thesection.\arabic{rifttheorem}}
\setcounter{riftproposition}{0}\renewcommand{\theriftproposition}{\thesection.\arabic{riftproposition}}
\setcounter{riftlemma}{0}\renewcommand{\theriftlemma}{\thesection.\arabic{riftlemma}}
\numberwithin{equation}{section}\setcounter{equation}{0}
\providecommand{\theHrifttheorem}{}\renewcommand{\theHrifttheorem}{\thesection.\arabic{rifttheorem}}
\providecommand{\theHriftproposition}{}\renewcommand{\theHriftproposition}{\thesection.\arabic{riftproposition}}
\providecommand{\theHriftlemma}{}\renewcommand{\theHriftlemma}{\thesection.\arabic{riftlemma}}

\subsection{Formal setting and conventions}\label{app:theory-setting}
A labeled transition is $\xi=(x,r)\in\Xi:=\mathbb R^{d_x}\times\mathbb R$ with $x=(s,a,s')$, and $\eta_c^y$ denotes its population law in reward class $y\in\{0,1\}$ of context $c\in\{A,B\}$. For $q\in(0,1/2)$, classes are the bottom and top $q$ fractions of rewards. At atoms on a threshold, independent uniform ranks break ties so that each class has probability $q$.

Write $\mathcal U=\mathbb R^{d_z}\times\mathbb R$ and $d_u=d_z+1$. The lift $E(x,r)=(e_\vartheta(x),r)$ and decoder $D(z,r)=(g_\vartheta(z),r)$ give $\mu_c^y=E_\#\eta_c^y$. The fixed normalization $S=S_B^{-1}\circ S_A$ uses the coordinatewise means and positive standard deviations of each full context distribution, including all reward classes; empirically, these statistics use all samples from that context. Thus $\bar\mu_A^y=S_\#\mu_A^y$. Squared latent matching is computed in these target coordinates, not before an additional anisotropic rescaling.

The Euclidean norm is used on both joint latent--reward and decoded transition--reward coordinates; the same symbol on matrices means the operator norm, with $\|\cdot\|_F$ the Frobenius norm. Write $\|f\|_{L^2(\mu)}=(\int\|f\|^2d\mu)^{1/2}$, and let $W_2$ denote quadratic Wasserstein distance. All laws entering these distances have finite second moments. Standard coupling and pushforward properties are used below \citep{villani2009optimal}.

The source coupling in Section~\ref{sec:theory} can be any fixed minimizer of
\[
\inf_{\pi\in\Pi(\bar\mu_A^0,\bar\mu_A^1)}
\mathbb E_{(X,Y)\sim\pi}\|Z_Y-Z_X\|^2,
\]
where $Z_X,Z_Y$ are the latent blocks. Its endpoints remain joint latent--reward points. The affine fit and residual therefore include the reward displacement. If the affine minimizer is not unique, all subsequent quantities refer to one fixed representative. Synthesis uses only the coupling's marginals, not its optimality, determinism, or an absolutely continuous class law.

The theorem, Theorem~\ref{thm:synthesis}, conditions on the learned representation, normalization, coupling, and field. The sampling and estimation results below use separate conditional-i.i.d. assumptions; they do not provide a generalization guarantee for jointly learned quantities.

\subsection{Synthesis, transplant, and sampling}\label{app:synthesis}
\begin{proof}[Proof of Theorem~\ref{thm:synthesis}]
Define the affine comparison map
\[
G_h=h\circ(\mathrm{id}+\delta_{\rm aff})\circ h^{-1},
\qquad \operatorname{Lip}(G_h)=\|H(I+W)H^{-1}\|=L_h.
\]
Then
\[
F-G_h=(\widehat\delta-\delta_{\rm aff})+\Gamma_h,
\qquad \|F-G_h\|_{L^2(\mu_B^0)}\le\varepsilon_{\rm est}+\varepsilon_{\rm conj}.
\]
For $(X,Y)\sim\pi_A$, the pair $(G_h(h(X)),h(Y))$ couples $(G_h)_\#h_\#\bar\mu_A^0$ and $h_\#\bar\mu_A^1$, with difference $H(X+\delta_{\rm aff}(X)-Y)$. Its root-mean-square cost is at most $\|H\|\varepsilon_{\rm aff}$. The triangle inequality and Lipschitz pushforward bound therefore give
\begin{equation}\label{eq:affine-transfer}
\begin{aligned}
W_2(F_\#\mu_B^0,\mu_B^1)
&\le \|F-G_h\|_{L^2(\mu_B^0)}
  +L_hW_2(\mu_B^0,h_\#\bar\mu_A^0)
  +\|H\|\varepsilon_{\rm aff}+\rho_1\\
&\le\varepsilon_{\rm est}+\varepsilon_{\rm conj}
  +\|H\|\varepsilon_{\rm aff}+L_h\rho_0+\rho_1.
\end{aligned}
\end{equation}
The nonlinear or nondeterministic affine-approximation error is paid only under the source coupling. The lifted decoder is $L_D=\max(1,L_g)$-Lipschitz. Coupling $D(E(\xi))$ to $\xi\sim\eta_B^1$ gives $W_2(D_\#\mu_B^1,\eta_B^1)\le\varepsilon_{\rm dec}$. Hence
\[
W_2(\nu,\eta_B^1)
\le L_DW_2(F_\#\mu_B^0,\mu_B^1)+\varepsilon_{\rm dec}
\le\varepsilon_{\rm syn}.
\]
An optimal coupling of the decoded laws gives the claimed root-mean-square interpretation.
\end{proof}

\paragraph{Context interpretation.}
For $h(u)=Hu+t$,
\[
\Gamma_h(u)=(W-HWH^{-1})u+HWH^{-1}t+(I-H)b.
\]
For a shift this equals $Wt$. For a rotation $H=O$, $t=0$, and a constant field $W=0$, it equals $(I-O)b$. Thus a constant improvement is invariant to a common translation but does not need to be invariant to a rotation. The result does not assume that normalization removes every residual class-relative or cross-coordinate change.

\begin{proof}[Proof of Proposition~\ref{prop:transplant}]
The encoder lift is $L_E=\max(1,L_e)$-Lipschitz, so
\[
W_2(D_\#\bar\mu_A^1,\eta_B^1)
\ge L_E^{-1}W_2((E\circ D)_\#\bar\mu_A^1,\mu_B^1).
\]
Since $(E\circ D)(z,r)=(e_\vartheta(g_\vartheta(z)),r)$, coupling by the same source-high point gives $W_2((E\circ D)_\#\bar\mu_A^1,\bar\mu_A^1)\le\varepsilon_{\rm cyc}$. The reverse triangle inequality proves \eqref{eq:transplant-bound}. Under Assumption~\ref{ass:transfer}, a second reverse triangle inequality yields
\[
W_2(\bar\mu_A^1,\mu_B^1)
\ge W_2(\bar\mu_A^1,h_\#\bar\mu_A^1)-\rho_1.
\]
For a shift, $(X,X+t)$ is a coupling with cost $\|t\|$, while every coupling satisfies $\mathbb E\|X-Y\|^2\ge\|\mathbb E(X-Y)\|^2=\|t\|^2$. Thus $W_2(\bar\mu_A^1,h_\#\bar\mu_A^1)=\|t\|$. Combining the lower bound with Theorem~\ref{thm:synthesis} gives the strict comparison.
\end{proof}

\paragraph{Finite synthetic samples.}\label{app:sampling}
Under the conditional-i.i.d. anchor assumption of Appendix~\ref{app:theory-setting}, let $\widehat\xi_i=D(F(U_i))$, $i=1,\ldots,m$, be the actual generated points and $\widehat\nu_m$ their empirical law. On an enlarged probability space they admit partners $\xi_i^\star\sim\eta_B^1$ such that the pairs $(\widehat\xi_i,\xi_i^\star)$ are i.i.d. and $\mathbb E\|\widehat\xi_i-\xi_i^\star\|^2\le\varepsilon_{\rm syn}^2$. With $\widehat\nu_m^\star$ their partners' empirical law,
\begin{equation}\label{eq:empirical-sampling}
\mathbb E W_2^2(\widehat\nu_m,\widehat\nu_m^\star)\le\varepsilon_{\rm syn}^2,
\qquad
\mathbb E W_2(\widehat\nu_m,\eta_B^1)
\le\varepsilon_{\rm syn}+\mathbb E W_2(\widehat\nu_m^\star,\eta_B^1).
\end{equation}
\begin{proof}
Disintegrate an optimal coupling of $(\nu,\eta_B^1)$ at each actual $\widehat\xi_i$, using independent auxiliary draws across indices to obtain $\xi_i^\star$. Euclidean spaces admit these conditional laws. The pairs have the claimed laws and squared cost. Matching pairs proves the first inequality; the triangle inequality and Jensen prove the second.
\end{proof}
If $\varepsilon_{\rm syn}=0$, then $\nu=\eta_B^1$. For a measurable region $\mathcal O$ fixed under the conditioning, at least one of the $m$ independent generated points hits $\mathcal O$ with probability $1-(1-\eta_B^1(\mathcal O))^m$. Thus synthesis can fill empirical coverage gaps without expanding the support of the target conditional law. These statements concern candidates before rejection and condition on the retained anchor sample.

\subsection{Estimating the field with latent-only minibatch matching}
\label{app:estimation}

This section analyzes the matching cost used by RIFT under an
independent-batch sampling model. The additional regularity and
sampling conditions below are needed only for this estimation
result, not for Theorem~\ref{thm:synthesis}.

\paragraph{Population conditions.}
Suppose the joint source classes are supported in
$\{\|u\|\le R_u\}$, where $R_u>0$. Put
$\widetilde U=(U,1)$ and define
\[
M_A
=
\mathbb E_{\bar\mu_A^0}
[\widetilde U\widetilde U^\top]
\succ0,
\qquad
M_B
=
\mathbb E_{\mu_B^0}
[\widetilde U\widetilde U^\top],
\]
\[
a_*=\lambda_{\min}(M_A),
\qquad
\kappa_M
=
\|M_B^{1/2}M_A^{-1/2}\|.
\]

Let $P_y=\bar\mu_{A,z}^y$ denote the source latent marginal for
class $y$. Suppose that
$T_z=\nabla\varphi_z$ pushes $P_0$ to $P_1$, where
$\varphi_z$ is globally convex and $L_T$-smooth, with $L_T>0$.
On the normalized source high class, assume
\[
R=\psi_A(Z)
\quad\text{almost surely},
\]
where $\psi_A$ is $L_\psi$-Lipschitz on its support.

\paragraph{Reference coupling.}
For this estimation analysis, specialize the population source
coupling used in Theorem~\ref{thm:synthesis} to
\[
\mathcal T_A(z,r)
=
\bigl(T_z(z),\psi_A(T_z(z))\bigr),
\qquad
\pi_A
=
(\mathrm{id},\mathcal T_A)_\#
\bar\mu_A^0.
\]
The lift $\mathcal T_A$ is needed only
$\bar\mu_A^0$-almost surely.

\paragraph{Estimation experiment.}
Conditionally on the fixed representation and normalization,
draw $N_A$ independent points from each source class, with the
two samples independent. Divide each sample into
\[
J=\frac{N_A}{k}
\]
disjoint batches of size $k$, where $k\ge1$ divides $N_A$.
Within each pair of batches, solve the latent assignment problem
exactly, using a fixed measurable tie rule and independent
batchwise randomness when the matching procedure is randomized.
Fit $\widehat\delta$ by ordinary least squares to all resulting
joint latent--reward displacements.

\paragraph{Minibatch-bias quantity.}
Let $\widehat P_k^y$ denote the empirical latent law of $k$
independent draws from $P_y$. Define the expected excess
matching cost and its associated bias certificate by
\begin{equation}\label{eq:batch-excess}
\operatorname{excess}_k
:=
\mathbb E
W_2^2(\widehat P_k^0,\widehat P_k^1)
-
W_2^2(P_0,P_1)
\ge0,
\qquad
B_k
:=
\sqrt{
(1+L_\psi^2)L_T
\operatorname{excess}_k
}.
\end{equation}

\begin{riftproposition}[Finite-sample field estimation]
\label{prop:estimation}
Under the setup above, there exist constants $C_1,C_2$,
depending only on $R_u$ and $a_*$, such that for every
$\eta\in(0,1)$ and
\[
J\ge C_1\log(2/\eta),
\]
with probability at least $1-\eta$,
\begin{equation}\label{eq:field-estimation}
\varepsilon_{\rm est}
\le
\kappa_M
\left[
B_k
+
C_2
\sqrt{\frac{\log(2/\eta)}{J}}
\right].
\end{equation}
\end{riftproposition}

\paragraph{Scope.}
The smooth latent transport, source-high reward graph,
positive-design condition, and independent-batch experiment are
sufficient conditions for Proposition~\ref{prop:estimation};
they are not required by Theorem~\ref{thm:synthesis}.
The smooth transport is assumed and does not need to exist for arbitrary
singular class laws. The condition $M_A\succ0$ can also fail,
for example when the source low-class reward is an affine
function of its latent coordinates. In such cases,
Theorem~\ref{thm:synthesis} still applies to a fixed affine
field and its actual estimation error, but
Proposition~\ref{prop:estimation} does not provide a certificate
for that error.

\begin{riftlemma}[Same-marginal excess-cost control]\label{lem:excess}
Let $P,Q$ have finite second moments. Suppose $T=\nabla\varphi$ pushes $P$ to $Q$, where $\varphi$ is convex and globally $L$-smooth, $L>0$. For every $\pi\in\Pi(P,Q)$,
\begin{equation}\label{eq:excess-endpoint}
\mathbb E_\pi\|Y-T(X)\|^2
\le L\left[\mathbb E_\pi\|X-Y\|^2-W_2^2(P,Q)\right].
\end{equation}
\end{riftlemma}
\begin{proof}
This is the transport-plan estimate of \citet[Proposition~3.1]{li2021quantitative}; the following gap argument gives the stated finite-moment version. For the ordinary conjugate, evaluating at $w=x+(y-T(x))/L$ and using smoothness gives
\[
\varphi(x)+\varphi^*(y)-\langle x,y\rangle\ge\frac{\|y-T(x)\|^2}{2L}.
\]
Global smoothness bounds $|\varphi(x)|$ quadratically and $\|T(x)\|$ linearly. Fenchel equality at $T(X)$ therefore makes $\varphi^*$ integrable under $Q=T_\#P$. The gap is nonnegative and zero on $(\mathrm{id},T)_\#P$, so this coupling minimizes quadratic cost. With fixed marginals, the integrated gap is half the excess cost in \eqref{eq:excess-endpoint}. Integration proves the claim.
\end{proof}

\begin{proof}[Proof of Proposition~\ref{prop:estimation}]
Let $\Pi_k$ be the law of a uniformly selected matched joint pair from one batch. A permutation uses every point once, so $\Pi_k\in\Pi(\bar\mu_A^0,\bar\mu_A^1)$; its latent cost is $\mathbb E W_2^2(\widehat P_k^0,\widehat P_k^1)$. This proves nonnegativity in \eqref{eq:batch-excess}. Let $\delta_k$ be the population affine fit under $\Pi_k$. Since $M_B\preceq\kappa_M^2M_A$, every affine $f$ satisfies $\|f\|_{L^2(\mu_B^0)}\le\kappa_M\|f\|_{L^2(\bar\mu_A^0)}$.

\emph{Bias.} Put $a=\delta_k-\delta_{\rm aff}$. The normal equations and common input marginal give
\[
\|a\|_{L^2(\bar\mu_A^0)}^2
=\mathbb E_{\Pi_k}\langle Y-\mathcal T_A(X),a(X)\rangle
\le\bigl(\mathbb E_{\Pi_k}\|Y-\mathcal T_A(X)\|^2\bigr)^{1/2}\|a\|_{L^2(\bar\mu_A^0)}.
\]
Both high endpoints lie on the source reward graph, so
\[
\|Y-\mathcal T_A(X)\|^2
\le (1+L_\psi^2)\|Z_Y-T_z(Z_X)\|^2.
\]
Apply Lemma~\ref{lem:excess} to the latent marginal of the mean population coupling $\Pi_k$, not to an individual empirical coupling with different marginals. It follows that $\|\delta_k-\delta_{\rm aff}\|_{L^2(\bar\mu_A^0)}\le B_k$.

\emph{Sampling.} Write $\delta_k(u)=\Theta_k\widetilde u$, $\widehat\delta(u)=\widehat\Theta\widetilde u$, and $R_1=\sqrt{R_u^2+1}$. The normal equations give $\Theta_k=\mathbb E_{\Pi_k}[(Y-X)\widetilde X^\top]M_A^{-1}$, so $\|\Theta_k\|\le2R_uR_1/a_*$. For batch $j$ set
\[
M_j=\frac1k\sum_{i\in j}\widetilde X_i\widetilde X_i^\top,
\qquad Z_j=\frac1k\sum_{i\in j}(Y_i-X_i-\Theta_k\widetilde X_i)\widetilde X_i^\top.
\]
These are independent across batches, with $\mathbb E M_j=M_A$, $\mathbb E Z_j=0$, $\|M_j\|_F\le R_1^2$, and $\|Z_j\|_F\le B:=2R_uR_1(1+R_1^2/a_*)$. For independent matrices $A_j$ with $\|A_j\|_F\le B_0$, independence and Jensen bound the expected centered-average Frobenius norm by $B_0/\sqrt J$. McDiarmid's inequality \citep{mcdiarmid1989bounded}, with per-batch sensitivity $2B_0/J$, yields
\[
\left\|\frac1J\sum_j(A_j-\mathbb E A_j)\right\|_F
\le\frac{B_0}{\sqrt J}\left(1+\sqrt{2\log(1/\zeta)}\right)
\]
with probability at least $1-\zeta$. Apply this to $M_j$ and $Z_j$ with $\zeta=\eta/2$ each. Put $\widehat M_A=J^{-1}\sum_jM_j$ and $\widehat Z=J^{-1}\sum_j Z_j$. For $J\ge C_1\log(2/\eta)$ the event gives $\|\widehat M_A-M_A\|\le a_*/2$, hence $\widehat M_A\succeq M_A/2$ and $\|\widehat M_A^{-1}\|\le2/a_*$. In particular,
\[
\|\widehat M_A^{-1}M_A^{1/2}\|^2
=\|\widehat M_A^{-1}M_A\widehat M_A^{-1}\|
\le2\|\widehat M_A^{-1}\|\le4/a_*.
\]
The empirical normal equations now give
\[
\widehat\Theta-\Theta_k=\widehat Z\widehat M_A^{-1},\qquad
\|\widehat\delta-\delta_k\|_{L^2(\bar\mu_A^0)}
=\|\widehat Z\widehat M_A^{-1}M_A^{1/2}\|_F
\le\frac2{\sqrt{a_*}}\|\widehat Z\|_F
\le C_2\sqrt{\frac{\log(2/\eta)}J}.
\]
For $c_*=1/\sqrt{\log2}+\sqrt2$, valid choices are $C_1=4c_*^2R_1^4/a_*^2$ and $C_2=2c_*B/\sqrt{a_*}$. Combining bias, sampling, and the affine source-to-target norm comparison proves \eqref{eq:field-estimation}.
\end{proof}

\paragraph{Rate and scope.}
The expected absolute quadratic-cost error of \citet[Theorem~2]{chizat2020faster}, rescaled from diameter one to a radius-$R_u$ ball, implies
\begin{equation}\label{eq:excess-rate}
\operatorname{excess}_k\le C_{d_z}R_u^2 r_{d_z}(k),\qquad
r_d(k)=\begin{cases}
k^{-1/2},&d<4,\\
k^{-1/2}\log(1+k),&d=4,\\
k^{-2/d},&d>4.
\end{cases}
\end{equation}
The expected excess is nonnegative and bounded by that absolute error. See \citet{manole2024sharp} for the broader smooth-cost analysis. Thus the high-dimensional bias certificate is of order $k^{-1/d_z}$, while the sampling term uses $J=N_A/k$ independent batches. Bias does not need to vanish as $N_A$ grows at fixed $k$ \citep{fatras2020learning}. Radius and moment conditioning also carry dimension dependence; the bound is not dimension-free.

Note that proposition~\ref{prop:estimation} is an idealized independent-batch result. Its assumptions—including the reward graph, fixed representation, and positive design condition—are sufficient conditions for the bound rather than empirical diagnostics of the reported experiments.

\subsection{Conditional target-return guarantee}\label{app:policy}
The following result connects synthesis accuracy to a specified fitted or analytical surrogate.

\begin{enumerate}[label=\textbf{(R\arabic*)},leftmargin=3.1em,itemsep=4pt]
\item \textbf{Target and data laws.}
The target has discount $\gamma\in[0,1)$, initial law $\rho$, bounded reward $r_B(s,a)$, and deterministic dynamics $S'=f_B(S,A)$. The reward and dynamics are respectively $L_r$- and $L_f$-Lipschitz on a domain containing the real, candidate synthetic, and comparison coordinates. Let $\eta_A,\eta_B$ be the full-context transition--reward laws underlying $\eta_c^y$. Genuine target tuples obey $R=r_B(S,A)$ and $S'=f_B(S,A)$. Source rewards are bounded but does not need to equal $r_B(S,A)$, and source dynamics may differ.

\item \textbf{Filtering and joint mixture.}
Fix a measurable acceptance event $K$ with $p_{\rm keep}=\nu(K)>0$, and let $\nu_K=\nu(\cdot\mid K)$. Retained actions are feasible at their states. Without filtering take $K=\Xi$ and $p_{\rm keep}=1$. For fixed mixture weights $\lambda,\theta\in[0,1]$, let
\begin{equation}\label{eq:training-law}
\eta_{\lambda,\theta}=(1-\lambda)\eta_A+\lambda(1-\theta)\eta_B+\lambda\theta\nu_K,
\qquad d_{\lambda,\theta}=\text{its state--action marginal}.
\end{equation}
For uniform sampling within the empirical target pool, the corresponding synthetic fraction is $\theta=m_{\rm keep}/(n_B+m_{\rm keep})$. The population training backup is
\begin{equation}\label{eq:mixture-backup}
(\mathcal B_{\lambda,\theta}V)(s,a)
=\mathbb E_{\eta_{\lambda,\theta}}[R+\gamma V(S')\mid S=s,A=a].
\end{equation}
It conditions the joint mixture: component weights does not need to remain constant after conditioning on $(s,a)$.

\item \textbf{Surrogate regularity and fitting.}
Let $\widehat{\mathcal M}$ be a fitted or analytical surrogate with the same spaces, discount, and initial law, bounded rewards, and state--action backup $\widehat{\mathcal B}$. Conditional laws, backups, policies, and values have fixed measurable definitions at all coordinates used below; extensions beyond observed support are part of the chosen model. For $\pi\in\{\widehat\pi,\pi_c\}$, its actual policy value $\widehat V^\pi=V_{\widehat{\mathcal M}}^\pi$ satisfies
\[
\|\widehat V^\pi\|_\infty\le V_{\max},\qquad
\operatorname{Lip}(\widehat V^\pi)\le L_V.
\]
The bounds also apply on the extensions used in the backups. Define
\begin{equation}\label{eq:policy-fit}
\varepsilon_{\rm fit}(\pi)=\mathbb E_{d_{\lambda,\theta}}
\left| (\widehat{\mathcal B}-\mathcal B_{\lambda,\theta})\widehat V^\pi(S,A)\right|.
\end{equation}

\item \textbf{Occupancy domination.}
For both policies, let
\begin{equation}\label{eq:occupancy}
d_B^\pi=(1-\gamma)\sum_{t=0}^\infty\gamma^t
\Pr_B^\pi((S_t,A_t)\in\cdot),
\qquad d_B^\pi\le C_\pi d_{\lambda,\theta}
\end{equation}
for a finite $C_\pi$. All displayed error expectations are finite. This is a population coverage condition; in continuous spaces a finite atomic empirical law cannot generally replace $d_{\lambda,\theta}$.
\end{enumerate}

\paragraph{Continuing error budget.}
Define source discrepancies in the decoded coordinates and reward units used by the policy learner:
\[
\varepsilon_{A,r}=\mathbb E_{\eta_A}|R-r_B(S,A)|,
\qquad \varepsilon_{A,f}=\mathbb E_{\eta_A}\|S'-f_B(S,A)\|.
\]
Write
\begin{equation}\label{eq:Ksyn}
K_{\rm syn}=\sqrt{1+(\gamma L_V)^2+(L_r+\gamma L_VL_f)^2},
\end{equation}
from the reward, next-state, and state--action blocks, and set
\begin{equation}\label{eq:data-budget}
\mathcal E_{\rm data}=(1-\lambda)(\varepsilon_{A,r}+\gamma L_V\varepsilon_{A,f})
+\lambda\theta K_{\rm syn}\frac{\varepsilon_{\rm syn}}{\sqrt{p_{\rm keep}}},
\qquad
\mathcal R_{\lambda,\theta}(\pi)=\frac{C_\pi}{1-\gamma}
[\varepsilon_{\rm fit}(\pi)+\mathcal E_{\rm data}].
\end{equation}
Here $\varepsilon_{\rm syn}$ is exactly \eqref{eq:synthesis-bound}. Source immediate-reward error has no factor $\gamma$; it vanishes under shared state--action rewards.

\subsubsection{Episodic definitions}\label{app:episodic}
For terminating tasks keep (R1)--(R4), with the following replacements. Let $\tau_B(s')\in\{0,1\}$ be the measurable target termination function, $\operatorname{term}_B(s,a)=\tau_B(f_B(s,a))$, and $\mathcal T_B=\{s':\tau_B(s')=1\}$. Genuine transitions have their own context's correct flag $d$; synthetic transitions use $d=\tau_B(s')$. The initial law is supported on target nonterminal states and the final transition's reward is included. The boundary of $\mathcal T_B$ is in the ambient Euclidean state space.

Attach the flags to the same mixture laws in \eqref{eq:training-law}. Replace the training backup by
\[
(\mathcal B_{\lambda,\theta}V)(s,a)
=\mathbb E_{\eta_{\lambda,\theta}}[R+\gamma(1-d)V(S')\mid S=s,A=a].
\]
The surrogate includes its own continuation flag, with (R3) still required. Hard termination does not need to preserve value regularity; Appendix~\ref{app:smooth-surrogate} supplies a concrete sufficient surrogate class. If $T$ is the target termination time, use the survival occupancy
\[
d_B^\pi=(1-\gamma)\sum_{t=0}^\infty\gamma^t
\Pr_B^\pi((S_t,A_t)\in\cdot,\ t<T)
\]
in (R4), and use flagged backups in $\varepsilon_{\rm fit}$.

Put $\varepsilon_{A,d}=\mathbb E_{\eta_A}|d-\operatorname{term}_B(S,A)|$ and $b_B(t)=\eta_B^1(\operatorname{dist}(S',\partial\mathcal T_B)\le t)$. For $\varepsilon>0$ and $p>0$, define
\begin{equation}\label{eq:omega}
\Omega_B(\varepsilon,p)=\min\left\{1,\frac1p\inf_{t>0}
\left[b_B(t)+\frac{\max\{1,L_f^2\}\varepsilon^2}{t^2}\right]\right\}.
\end{equation}
Set $\Omega_B(0,p)=0$; distance to an empty boundary is $+\infty$. For fixed $p$, the positive-error limit is $\min\{1,\eta_B^1(S'\in\partial\mathcal T_B)/p\}$, so it tends to zero when the boundary has zero target-high mass. Exact partners still have identical flags at zero error, including on the boundary.

In the episodic version, add
\begin{equation}\label{eq:termination-budget}
\mathcal E_{\rm term}=\gamma V_{\max}
\left[(1-\lambda)\varepsilon_{A,d}
+\lambda\theta\Omega_B(\varepsilon_{\rm syn},p_{\rm keep})\right]
\end{equation}
to $\mathcal E_{\rm data}$ in \eqref{eq:data-budget}, and use the resulting $\mathcal R_{\lambda,\theta}$. For continuing tasks all flags and $\mathcal E_{\rm term}$ are zero and $T=\infty$.

\begin{rifttheorem}[Synthesis-to-return comparison]\label{thm:policy}
Assume (R1)--(R4), using the continuing or episodic definitions, and the decoded coupling bound in Theorem~\ref{thm:synthesis}. Then for $\pi\in\{\widehat\pi,\pi_c\}$,
\begin{equation}\label{eq:return-fidelity}
|J_B(\pi)-J_{\widehat{\mathcal M}}(\pi)|\le\mathcal R_{\lambda,\theta}(\pi).
\end{equation}
If $J_{\widehat{\mathcal M}}(\widehat\pi)\ge J_{\widehat{\mathcal M}}(\pi_c)-\varepsilon_{\rm learn}$ for $\varepsilon_{\rm learn}\ge0$, then
\begin{equation}\label{eq:policy-comparison}
J_B(\widehat\pi)\ge J_B(\pi_c)-\varepsilon_{\rm learn}
-\mathcal R_{\lambda,\theta}(\widehat\pi)-\mathcal R_{\lambda,\theta}(\pi_c).
\end{equation}
The result holds for each realized surrogate and policies satisfying these conditions; it requires neither comparator optimality nor nested policy classes.
\end{rifttheorem}
Subtracting any baseline return $J_B(\pi_0)$ gives the corresponding improvement comparison without a coverage assumption on $\pi_0$. Strict gain follows only if the comparator's advantage over that baseline exceeds the error budget. No positive comparator advantage, IQL-specific rate, or data-driven certificate of these population quantities is supplied by the theorem.

\begin{riftlemma}[Uniform training-law consistency]\label{lem:training-consistency}
Under (R1)--(R2) and the decoded coupling bound, every $V$ with $\|V\|_\infty\le V_{\max}$ and $\operatorname{Lip}(V)\le L_V$ satisfies
\[
\mathbb E_{\eta_{\lambda,\theta}}|b_V|\le\mathcal E_{\rm data},
\quad b_V(\xi,d)=r+\gamma(1-d)V(s')-(\mathcal B_BV)(s,a),
\]
where $(\mathcal B_BV)(s,a)=r_B(s,a)+\gamma(1-\operatorname{term}_B(s,a))V(f_B(s,a))$. Use zero flags and the continuing budget in continuing tasks.
\end{riftlemma}
\begin{proof}
For any flagged transition, adding and subtracting $(1-d)V(f_B(s,a))$ gives
\begin{equation}\label{eq:pointwise-backup}
|b_V(\xi,d)|\le |r-r_B(s,a)|+\gamma L_V\|s'-f_B(s,a)\|
+\gamma V_{\max}|d-\operatorname{term}_B(s,a)|.
\end{equation}
Couple $\widehat\xi\sim\nu$ to $\xi^\star\sim\eta_B^1$ as in Theorem~\ref{thm:synthesis}, and put $Z=\|\widehat\xi-\xi^\star\|$, so $\mathbb E Z^2\le\varepsilon_{\rm syn}^2$. The phantom is target-consistent. Lipschitz continuity and Cauchy--Schwarz on the reward, next-state, and state--action blocks bound the first two terms by $K_{\rm syn}Z$. Conditioning on acceptance gives mean at most
\[
K_{\rm syn}\frac{\mathbb E[Z\mathbf1_K]}{p_{\rm keep}}
\le K_{\rm syn}\frac{\varepsilon_{\rm syn}}{\sqrt{p_{\rm keep}}}.
\]
For termination put $M_\star=\operatorname{dist}(s^{\star\prime},\partial\mathcal T_B)$ and $a=\max(1,L_f)$. Both $\widehat s'$ and $f_B(\widehat s,\widehat a)$ lie within $aZ$ of $s^{\star\prime}$. If $M_\star>aZ$, that ball avoids the boundary, so their flags agree. Splitting on $Z>t/a$ and using Markov's inequality gives, for each $t>0$,
\[
p_{\rm keep}\mathbb E_{\nu_K}|d-\operatorname{term}_B(S,A)|
\le b_B(t)+a^2\varepsilon_{\rm syn}^2/t^2.
\]
Taking the infimum and capping the conditional probability at one yields \eqref{eq:omega} for positive error. For zero error the partners coincide almost surely, so flags agree even on the boundary. Thus
\[
\mathbb E_{\nu_K}|b_V|\le K_{\rm syn}\frac{\varepsilon_{\rm syn}}{\sqrt{p_{\rm keep}}}
+\gamma V_{\max}\Omega_B(\varepsilon_{\rm syn},p_{\rm keep}).
\]
On target data $b_V=0$; on source data \eqref{eq:pointwise-backup} gives $\mathbb E_{\eta_A}|b_V|\le\varepsilon_{A,r}+\gamma L_V\varepsilon_{A,f}+\gamma V_{\max}\varepsilon_{A,d}$. Expanding the mixture proves the claim. All flag terms vanish in the continuing case. This bounds retained-sample consistency, not the filtered law's Wasserstein distance.
\end{proof}

\begin{proof}[Proof of Theorem~\ref{thm:policy}]
\emph{Covered Bellman error.} For either policy write $e_\pi=(\widehat{\mathcal B}-\mathcal B_B)\widehat V^\pi$. Conditional Jensen, Lemma~\ref{lem:training-consistency}, the fitting definition, and occupancy domination give
\begin{equation}\label{eq:onpolicy-budget}
\mathbb E_{d_B^\pi}|e_\pi|
\le C_\pi\mathbb E_{d_{\lambda,\theta}}|e_\pi|
\le C_\pi[\varepsilon_{\rm fit}(\pi)+\mathcal E_{\rm data}].
\end{equation}
The consistency estimate is uniform over bounded Lipschitz $V$, so no independence of the learned surrogate is needed.

\emph{Trajectory telescoping.} We use the same-policy Bellman-residual identity (e.g., \citealp[Lemma~1]{xie2020qstar}), retaining survival indicators for episodic tasks. Let $I_t=\mathbf1_{\{t<T\}}$ along a target trajectory, extended arbitrarily after termination, and let $r_t=r_B(S_t,A_t)$. Then $I_t(1-\operatorname{term}_B(S_t,A_t))=I_{t+1}$. The surrogate policy Bellman equation, averaged over the action at each live state, yields
\[
\sum_{t=0}^{n-1}\gamma^t\mathbb E_B^\pi[I_te_\pi(S_t,A_t)]
=J_{\widehat{\mathcal M}}(\pi)-\mathbb E_B^\pi\sum_{t=0}^{n-1}\gamma^t I_tr_t
-\gamma^n\mathbb E_B^\pi[I_n\widehat V^\pi(S_n)].
\]
The final term is bounded by $\gamma^n V_{\max}$. Bounded rewards and values allow $n\to\infty$, giving
\[
J_{\widehat{\mathcal M}}(\pi)-J_B(\pi)
=\frac1{1-\gamma}\mathbb E_{d_B^\pi}e_\pi.
\]
Together with \eqref{eq:onpolicy-budget}, this proves \eqref{eq:return-fidelity}. The learned-policy condition then gives
\[
J_B(\widehat\pi)\ge J_{\widehat{\mathcal M}}(\widehat\pi)-\mathcal R_{\lambda,\theta}(\widehat\pi)
\ge J_B(\pi_c)-\varepsilon_{\rm learn}
-\mathcal R_{\lambda,\theta}(\widehat\pi)-\mathcal R_{\lambda,\theta}(\pi_c),
\]
which is \eqref{eq:policy-comparison}. Setting $T=\infty$ and all flags to zero proves the continuing case with the same argument.
\end{proof}

\paragraph{Interpretation.}
Filtering and termination enter through the displayed consistency penalties, while source reward and dynamics mismatch can remain even with perfect synthesis. The theorem leaves the source-data effects on coverage, fitting, and optimization explicit, so it does not select an optimal \(\lambda\). The quantities \(\varepsilon_{\rm fit}\) and \(\varepsilon_{\rm learn}\) are theoretical error terms rather than observed training losses.

\subsection{A regular analytical surrogate}\label{app:smooth-surrogate}
Hard continuation flags can make exact surrogate values discontinuous even at nonterminal states. The correction $\Omega_B$ controls flag disagreement, not those value jumps. The following construction verifies (R3) for a common analytical reference. It is a finite-outcome model of the kernel-interpolation type \citep{ormoneit2002kernel}.

Fix outcomes $(y_i,r_i,d_i)$, $i=1,\ldots,n_*$, with $|r_i|\le R_*$, $R_*>0$, $d_i\in\{0,1\}$, and $y_i$ in the state domain, including required extensions. Choose nonnegative smooth weights with
\[
\sum_iw_i(u)=1,\qquad
\sum_i|w_i(u)-w_i(v)|\le L_w\|u-v\|,\quad L_w<\infty.
\]
At $(s,a)$, select $i$ with probability $w_i(s,a)$, receive $r_i$, and stop if $d_i=1$; otherwise continue at $y_i$. The resulting stochastic surrogate has the same discount and initial law, and
\[
(\widehat{\mathcal B}V)(s,a)
=\sum_iw_i(s,a)[r_i+\gamma(1-d_i)V(y_i)].
\]
Stopping uses the sampled flag, not an additional hard stopping rule imposed by the target terminal set. Both continuation probability and the weighted continuation outcomes vary smoothly.

\begin{riftlemma}[Value regularity of the interpolated surrogate]\label{lem:smooth-surrogate}
For any deterministic policy with $\operatorname{Lip}(\pi)\le L_\pi$ on this domain, the exact surrogate value satisfies
\[
\|\widehat V^\pi\|_\infty\le V_*:=\frac{R_*}{1-\gamma},
\qquad \operatorname{Lip}(\widehat V^\pi)\le V_*L_w\sqrt{1+L_\pi^2}.
\]
No contraction of physical dynamics in the Euclidean state metric is assumed.
\end{riftlemma}
\begin{proof}
The policy Bellman operator is a sup-norm contraction with modulus at most $\gamma$, and bounded rewards give $V_*$. At its fixed point, $\widehat V^\pi(s)=\sum_iw_i(s,\pi(s))b_i^\pi$, where $b_i^\pi=r_i+\gamma(1-d_i)\widehat V^\pi(y_i)$ is independent of the current state and $|b_i^\pi|\le R_*+\gamma V_*=V_*$. Therefore
\[
|\widehat V^\pi(s)-\widehat V^\pi(t)|
\le V_*\sum_i|w_i(s,\pi(s))-w_i(t,\pi(t))|
\le V_*L_w\sqrt{1+L_\pi^2}\|s-t\|.
\]
\end{proof}
For deterministic Lipschitz $\widehat\pi,\pi_c$, one common surrogate satisfies (R3) with $V_{\max}=V_*$ and $L_V=V_*L_w\max_{\pi\in\{\widehat\pi,\pi_c\}}\sqrt{1+L_\pi^2}$. The result holds for every realized outcome set and weight function satisfying these conditions, including fitted choices.

\paragraph{Fitting cost and scope.}
For any measurable conditional coupling $\Gamma$ of training outcomes $(R,S',d)$ and surrogate outcomes $(\widetilde R,\widetilde S',\widetilde d)$ at the same $(S,A)$, with expectation also over $d_{\lambda,\theta}$, conditional Jensen and the add-and-subtract argument in \eqref{eq:pointwise-backup} give
\[
\varepsilon_{\rm fit}(\pi)\le\mathbb E_\Gamma
\left[|R-\widetilde R|+\gamma L_V\|S'-\widetilde S'\|
+\gamma V_{\max}|d-\widetilde d|\right].
\]
Smoothing discrepancy enters $\varepsilon_{\rm fit}$. The target-flag and filtering penalties, including $\Omega_B$, still apply. Steeper weights can enlarge $L_V$, while broader averaging can enlarge fitting error; neither is asserted small. Coverage and the learning comparison must hold for this same surrogate. This construction gives a sufficient condition for value regularity for the surrogate class considered here.

\section{Environment Setting}
\label{app:datasets}
We construct the same environment settings as the ODRL benchmark~\citep{lyu2024odrl} and use the ODRL offline datasets
directly. This matches the benchmark protocol used by recent cross-domain offline RL methods, including OTDF, STC, and DmC~\citep{lyu2025otdf,yan2026stc,van2025dmc}, allowing for comparison under the same settings.

\subsection{Source and Target Datasets}

We consider four MuJoCo locomotion tasks: Ant, HalfCheetah, Hopper, and Walker2d. The source datasets are the corresponding D4RL datasets distributed with ODRL. We consider medium (\texttt{med}), medium-replay (\texttt{med-r}), and medium-expert (\texttt{med-e}) source datasets. The full source dataset is used in each experiment and contains approximately one million transitions, depending on the D4RL dataset~\citep{fu2020d4rl}.

For the target domain, we use the limited offline datasets provided by ODRL. ODRL provides medium and expert target datasets containing $5,000$ transitions. To construct the medium-expert dataset, we follow the OTDF and STC protocol of combining two whole trajectories from the medium dataset with three from the expert dataset, to get a total of $5{,}000$ transitions. We use the provided target datasets directly rather than recollecting trajectories. For each experimental setting, the same target dataset is held fixed across all five policy training seeds.

\subsection{Target Dynamics Shifts}

The source environment uses the default MuJoCo dynamics. The target environment modifies these dynamics according to the ODRL benchmark. The descriptions below follow its released XML assets~\citep{odrl_code}. We study gravity, friction, and morphology shifts.

\paragraph{Gravity shift.}
For the gravity shift, the magnitude of the gravitational acceleration is reduced by a factor of $0.5$ while preserving its direction. In the MuJoCo XML, this corresponds to changing

\begin{verbatim}
<option gravity="0 0 -9.81" ... />
\end{verbatim}

to

\begin{verbatim}
<option gravity="0 0 -4.905" ... />
\end{verbatim}

for the target environment.

\paragraph{Friction shift.}
ODRL halves explicitly specified geometry friction values. In Hopper, the torso, thigh, and leg entries change from \texttt{friction="0.9"} to \texttt{friction="0.45"}, and the foot entry changes from $2.0$ to $1.0$. The other tasks halve their explicit tuples and geometry-specific overrides; omitted coefficients retain inherited defaults.

\paragraph{Morphology shift.}
For the morphology shift, we use ODRL's medium-level change to one body part of each robot. Unlike the gravity and friction shifts, this alters model geometry rather than gravity or friction coefficients:
\begin{itemize}
\item \textbf{HalfCheetah (thigh):} the back and front thigh capsules are shortened from approximate half-lengths $0.145$ and $0.133$ to $0.057$ and $0.053$, with the shin and foot bodies repositioned accordingly.
\item \textbf{Hopper (foot):} the foot capsule shrinks from radius $0.06$ and half-length $0.195$ to $0.036$ and $0.117$, and the foot body moves downward by $0.10$.
\item \textbf{Walker2d (leg):} on both legs, the thigh capsule lengthens from half-length $0.225$ to $0.35$ and the shin shortens from $0.25$ to $0.125$, with the foot bodies moved downward by $0.10$.
\item \textbf{Ant (legs):} the four ankle capsules are shortened from approximate half-length $0.283$ to $0.141$ by halving their endpoint offsets; the ankle-body origins are unchanged.
\end{itemize}
For example, Hopper's global-coordinate XML changes the foot capsule from

\begin{verbatim}
<geom name="foot_geom" size="0.06"
      fromto="-0.13 0 0.1 0.26 0 0.1" ... />
\end{verbatim}

to

\begin{verbatim}
<geom name="foot_geom" size="0.036"
      fromto="-0.078 0 0.1 0.156 0 0.1" ... />
\end{verbatim}

in the target environment.

\subsection{Evaluation}

Policies are evaluated only in the corresponding target environment. Following ODRL~\citep{lyu2024odrl,odrl_code}, the evaluation statistic $G$ is the mean undiscounted episodic return, distinct from the discounted $J_{\mathcal M}$ used in the theory. The D4RL-style normalized score is

\[
\mathrm{NS}
=
100 \cdot 
\frac{G-G_{\mathrm{random}}}
     {G_{\mathrm{expert}}-G_{\mathrm{random}}},
\]

where $G_{\mathrm{random}}$ and $G_{\mathrm{expert}}$ are the benchmark reference scores for the corresponding task and shift. Unless otherwise stated, reported results use the final policy checkpoint and average over five policy seeds. The target dataset is fixed across those seeds. For RIFT, each seed refits the VAE, recomputes the OT couplings and field, and retrains the policy, so variation covers full-pipeline training rather than independently sampled target datasets. Appendix~\ref{app:target-subsets} separately studies variance from differences in target-data.

\section{Algorithmic Implementation}
\label{app:hyperparams}

We follow the implementation settings used in the ODRL benchmark along with their corresponding baseline configurations. We first verify our IQL implementation against the reference results reported by STC, and then list the hyperparameters used for all methods.

\subsection{IQL Replication}

Table~\ref{tab:iql-reference} compares our IQL implementation with the IQL results reported by STC. We reproduce both the pooled source--target setting and the target-only setting on the eight medium-to-medium gravity and morphology tasks. The pooled aggregate is close to STC (234.0 vs. 232.4), while the target-only aggregate is higher (312.4 vs. 282.0); individual task scores differ.

\begin{table}[t]
\centering
\caption{IQL reference results across eight medium-to-medium gravity and morphology settings. Values report the mean $\pm$ standard deviation over five seeds, with the highest mean bolded.}
\label{tab:iql-reference}
\setlength{\tabcolsep}{4pt}
\resizebox{\textwidth}{!}{%
\begin{tabular}{llccccc}
\toprule
& & \multicolumn{2}{c}{IQL (reported by STC)}
& \multicolumn{2}{c}{IQL (ours)}
& \\
\cmidrule(lr){3-4}\cmidrule(lr){5-6}
Shift & Environment
& $\mathcal D_A\cup\mathcal D_B$ & $\mathcal D_B$
& $\mathcal D_A\cup\mathcal D_B$ & $\mathcal D_B$
& \textbf{RIFT} \\
\midrule
Gravity & Ant
& \score{10.2}{1.8} & \score{32.0}{7.6}
& \score{10.1}{1.4} & \score{29.8}{4.9}
& \bestscore{40.8}{10.3} \\
& HalfCheetah
& \score{39.6}{3.3} & \score{0.6}{0.5}
& \score{42.0}{2.3} & \score{2.6}{0.6}
& \bestscore{47.3}{6.6} \\
& Hopper
& \score{11.2}{1.1} & \score{46.2}{3.7}
& \score{14.1}{4.8} & \score{52.7}{12.5}
& \bestscore{60.4}{7.3} \\
& Walker2d
& \score{28.1}{12.9} & \score{28.5}{7.4}
& \score{30.6}{4.3} & \score{26.7}{4.5}
& \bestscore{46.5}{5.5} \\
\midrule
Morphology & Ant
& \score{71.4}{2.4} & \score{76.8}{2.8}
& \score{65.5}{2.0} & \bestscore{76.9}{0.3}
& \score{76.2}{2.4} \\
& HalfCheetah
& \score{24.5}{2.4} & \score{12.3}{3.7}
& \score{22.8}{2.4} & \score{19.0}{6.6}
& \bestscore{57.9}{4.4} \\
& Hopper
& \score{15.9}{6.8} & \score{42.5}{17.3}
& \score{18.1}{7.1} & \score{55.5}{6.4}
& \bestscore{69.4}{13.2} \\
& Walker2d
& \score{31.5}{8.6} & \score{43.1}{11.5}
& \score{30.8}{9.2} & \score{49.2}{5.1}
& \bestscore{52.8}{8.2} \\
\midrule
\multicolumn{2}{l}{Total (displayed means)}
& 232.4 & 282.0 & 234.0 & 312.4 & \textbf{451.3} \\
\bottomrule
\end{tabular}%
}
\end{table}

\subsection{Hyperparameter Setup}
We include the baseline algorithms and hyperparameter settings as reported in STC~\citep{yan2026stc} and DmC~\citep{van2025dmc}.

\paragraph{IQL.}
IQL~\citep{kostrikov2022offline} learns a value function using expectile regression,
\[
\mathcal L_V
=
\mathbb E_{(s,a)\sim\mathcal D}
\left[
L_2^\tau\!\left(Q(s,a)-V(s)\right)
\right],
\]
where $L_2^\tau(u)=|\tau-\mathbf 1_{\{u<0\}}|u^2$. Its Q-function uses temporal-difference regression,
\[
\mathcal L_Q
=
\mathbb E_{(s,a,r,s',d)\sim\mathcal D}
\left[
\bigl(r+\gamma(1-d)V(s')-Q(s,a)\bigr)^2
\right].
\]
Here $d$ denotes environmental termination in the displayed discounted backup, not time-limit truncation by itself. The policy is trained by advantage-weighted behavior cloning,
\[
\mathcal L_\pi
=
-\mathbb E_{(s,a)\sim\mathcal D}
\left[
\exp\!\left(\beta_{\mathrm{IQL}}
(Q(s,a)-V(s))\right)
\log\pi(a\mid s)
\right].
\]

\paragraph{DARA.}
DARA~\citep{liu2022dara} trains domain classifiers to estimate the dynamics discrepancy between source and target transitions. The resulting discrepancy $\Delta r$ is used to modify the source reward as
\[
\hat r = r-\alpha\Delta r,
\]
with the reward penalty clipped to $[-10,10]$. The corrected source transitions are then used for downstream IQL training.

\paragraph{BOSA.}
BOSA~\citep{liu2024bosa} models target-domain support using a CVAE and restricts value learning to transitions supported by the estimated target dynamics. Its critic objective takes the
form
\[
\mathcal L_{\mathrm{BOSA}}
=
\mathbb E_{\mathcal D_{\mathrm{mix}}}
\!\left[
\delta(Q)\,
\mathbf 1\!\left(
\log\hat P_{\mathrm{tar}}(s'\mid s,a)>\epsilon'_{\mathrm{th}}
\right)
\right]
+
\mathbb E_{(s,a)\sim\mathcal D_A}[Q(s,a)],
\]
where $\delta(Q)$ denotes squared TD error, and $\epsilon'_{\mathrm{th}}$ is a log-likelihood threshold. The dynamics ensemble uses its minimum log-likelihood for this test. BOSA additionally constrains policy updates to behavior support in the pooled data~\citep{odrl_code}.

\paragraph{IGDF.}
IGDF~\citep{wen2024igdf} learns a contrastive representation that assigns each transition a score $h(s,a,s')$ reflecting source--target dynamics mismatch. Target samples are always used for value learning, while source samples are selected according to this score:
\[
\mathcal L_Q
=
\frac{1}{2}\mathbb E_{\mathcal D_B}
\!\left[(Q-\mathcal TQ)^2\right]
+
\frac{\alpha}{2}
\mathbb E_{\mathcal D_A}
\!\left[
\mathbf 1\!\left(h>h_{\xi\%}\right)
(Q-\mathcal TQ)^2
\right].
\]
The threshold $h_{\xi\%}$ determines the retained fraction of source transitions. 

\paragraph{SRPO.}
SRPO~\citep{xue2023srpo} estimates stationary-state information using a discriminator trained from high- and low-value transitions. The discriminator is then used to modify the reward,
\[
\hat r_{\mathrm{SRPO}}
=
r+\lambda
\frac{D_\delta(s)}
     {1-D_\delta(s)},
\]
and policy learning proceeds using the modified rewards.

\paragraph{OTDF.}
OTDF~\citep{lyu2025otdf} uses optimal transport to measure the dynamics discrepancy between source and target transitions. Given the optimal coupling $\mu^\star$, it assigns each source transition
$u_t$ the deviation score
\[
d(u_t)
=
-\sum_{t'}
C(u_t,u_{t'})\,\mu^\star_{t,t'}.
\]
In its practical implementation, OTDF computes the coupling using entropy-regularized optimal transport with the Sinkhorn algorithm and normalizes the resulting deviation scores. Source transitions are then selected according to the top-$\xi$ scores and weighted in the TD objective according to their target-domain compatibility. OTDF additionally trains a CVAE on the target dataset and uses it to regularize the learned policy toward the support of the target behavior policy.

\paragraph{STC.}
STC~\citep{yan2026stc} mitigates the dynamics gap by correcting source transitions toward the target domain. It first pretrains an inverse policy model, forward dynamics model, and reward model on the target dataset for $50{,}000$ steps. For each source transition, STC corrects the action using the inverse policy model and estimates the corrected reward using a first-order Taylor approximation. The corrected transition is retained only when its estimated target-domain dynamics discrepancy satisfies the correction criterion controlled by threshold $\lambda$. The value function is then trained with TD regression, while the policy uses a Q-weighted behavior-cloning objective with coefficient $\beta$.

\paragraph{DmC.}
DmC~\citep{van2025dmc} estimates source--target dynamics proximity using a $k$-nearest-neighbor score over transition tuples $(s,a,s')$. The resulting score is used both to select and weight source transitions during value learning and to condition a diffusion model that generates additional source samples closer to the target domain. DmC further applies a target-policy regularization term through a conditional VAE to reduce bias toward the source dataset. We include the DmC comparison in Appendix~\ref{app:DMC-results}.

\paragraph{IQL*.}
IQL* uses the same IQL implementation and hyperparameters as above, but follows the same data-mixing protocol used by RIFT, modifying the batch target-to-source ratio.

\paragraph{RIFT.}
We train a shared VAE using balanced source--target minibatches and add the additional reward-conditioned regularization from Section~\ref{sec:alg} to the loss:
\[
\mathcal L_{\mathrm{class}}
=
\eta\,
\mathbb E_{c\in\{A,B\}}
\mathbb E_{y\in\{0,1\}}
\mathbb E_{x\sim\mathcal D_c^y}
\left[
\mathrm{KL}\!\left(
q_\vartheta(z_0\mid x)
\,\|\, 
\mathcal N\!\big((2y-1)c_q,s_q^2\big)
\right)
\right].
\]

Here $q_\vartheta$ is the encoder posterior and $\eta$ weights the class penalty. We use mini-batch optimal transport with squared Euclidean cost to pair low- and high-reward source samples. Their joint latent--reward displacements are fit with an affine model using ordinary least squares; matching itself uses only squared latent distance. The learned field is then applied to low-reward target anchors to generate synthetic transitions. We similarly modify the batch target-to-source ratio.

\paragraph{Specifications.}

We summarize the hyperparameter settings for all methods in
Table~\ref{tab:all-hparams}. For IQL, DARA, and BOSA, we use the
fixed configurations reported by STC~\citep{yan2026stc}. For SRPO, the
reward coefficient is selected from $\{0.1,0.3\}$. For IGDF, the
data selection ratio is fixed to $75\%$, while the representation
dimension is selected from $\{16,64\}$. For OTDF, we follow the
reported implementation settings, using a fixed data selection
ratio of $80\%$ and a task-dependent policy coefficient in
$\{0.1,0.5\}$. For STC, the reward gradient coefficient is fixed
to $0.5$, while the correction threshold is selected from
$\{1.0,5.0\}$ and the Q-weighted loss coefficient from
$\{0.5,5.0\}$, with the best configuration reported for each
environment. For RIFT, we fix the representation and field-fitting hyperparameters across tasks and evaluate two downstream training configurations, $\lambda\in\{0.5,1\}$.

\clearpage
\begingroup
\small
\setlength{\tabcolsep}{5pt}
\begin{longtable}{lll}
\caption{Reported hyperparameter settings: shared IQL settings and method-specific configurations.}
\label{tab:all-hparams}\\
\toprule
Method & Hyperparameter & Value \\
\midrule

\multicolumn{3}{l}{\textit{Shared IQL backbone}} \\
& Actor network & $(256,256)$ \\
& Critic network & $(256,256)$ \\
& Learning rate & $3\times10^{-4}$ \\
& Optimizer & Adam \\
& Discount factor & $0.99$ \\
& Nonlinearity & ReLU \\
& Target update rate & $5\times10^{-3}$ \\
& Source-domain batch size & $128$ \\
& Target-domain batch size & $128$ \\
& Temperature coefficient & $0.2$ \\
& Maximum log std & $2$ \\
& Minimum log std & $-20$ \\
& Inverse temperature $\beta_{\mathrm{IQL}}$ & $3.0$ \\
& Expectile $\tau$ & $0.7$ \\
\midrule

DARA
& Classifier network & $(256,256)$ \\
& Reward penalty coefficient & $0.1$ \\
\midrule

BOSA
& Policy regularization coefficient & $0.1$ \\
& Transition coefficient & $0.1$ \\
& Threshold parameters $\epsilon,\epsilon'$ & $\log(0.01)$ \\
& Value weight & $0.1$ \\
& CVAE ensemble size & $5$ \\
\midrule

SRPO
& Discriminator network & $(256,256)$ \\
& Data selection ratio & $0.5$ \\
& Reward coefficient & $\{0.1,0.3\}$ \\
\midrule

IGDF
& Representation dimension & $\{16,64\}$ \\
& Contrastive encoder network & $(256,256)$ \\
& Encoder pretraining steps & $7{,}000$ \\
& Importance coefficient & $1.0$ \\
& Data selection ratio & $75\%$ \\
\midrule

OTDF
& CVAE training steps & $10{,}000$ \\
& CVAE learning rate & $10^{-3}$ \\
& Number of sampled latent variables & $10$ \\
& Gaussian standard deviation & $\sqrt{0.1}$ \\
& Cost function & cosine \\
& Data selection ratio & $80\%$ \\
& Policy coefficient & $\{0.1,0.5\}$ \\
\midrule

STC
& Correction threshold $\lambda$ & $\{1.0,5.0\}$ \\
& Reward gradient coefficient $\alpha$ & $0.5$ \\
& Q-weighted loss coefficient $\beta$ & $\{0.5,5.0\}$ \\
\midrule

DmC
& $k$-nearest neighbor & $5$ \\
& Data selection ratio & $50\%$ \\
& Policy regularization coefficient & $0.1$ \\
& Guided-sampling score threshold & $90\%$ \\
\midrule

IQL*
& Data-mixing coefficient $\lambda$ & $\{0.5,1\}$ \\
\midrule

RIFT
& Reward quantile $q$ & $0.2$ \\
& VAE latent dimension & $16$ \\
& VAE hidden layers & $(256,256)$ \\
& VAE learning rate / batch / steps & $10^{-3}$ / $256$ / $100{,}000$ \\
& OT mini-batch size & $2{,}000$ per side, $8$ problems \\
& Data-mixing coefficient $\lambda$ & $\{0.5,1\}$ \\
\bottomrule
\end{longtable}
\endgroup

\section{Additional Results}
\label{app:results}
We provide the detailed tables summarized in the main text and additional results.

\subsection{Target-Quality Results}\label{app:target-quality}
Table~\ref{tab:target-quality} gives the complete target-quality comparison summarized in Section~\ref{sec:main-results}.

\begin{table}[!htbp]
\centering
\caption{Performance under gravity and morphology shifts when the source dataset is fixed to medium quality and target-data quality varies. Values report the mean $\pm$ standard deviation over five seeds, with the highest mean bolded.}
\label{tab:target-quality}
\setlength{\tabcolsep}{3pt}
\resizebox{\textwidth}{!}{%
\begin{tabular}{lll*{8}{c}}
\toprule
Type & Source & Target
& IQL & DARA & BOSA & SRPO & IGDF & OTDF & STC
& \textbf{RIFT (ours)} \\
\midrule
Gravity & half-med & med
& \score{39.6}{3.3} & \score{41.2}{3.9} & \score{38.9}{4.0}
& \score{36.9}{4.5} & \score{36.6}{5.5} & \score{40.7}{7.7}
& \score{42.4}{5.3} & \bestscore{47.3}{6.6} \\
& half-med & med-e
& \score{39.6}{3.7} & \score{40.7}{2.8} & \score{40.4}{3.0}
& \score{40.7}{2.3} & \score{38.7}{6.2} & \score{28.6}{3.2}
& \score{37.3}{1.5} & \bestscore{46.2}{3.0} \\
& half-med & expert
& \score{42.4}{3.8} & \score{39.8}{4.4} & \score{40.5}{3.9}
& \score{39.4}{1.6} & \score{39.6}{4.6} & \score{36.1}{5.3}
& \score{43.1}{5.7} & \bestscore{45.5}{2.6} \\
& hopp-med & med
& \score{11.2}{1.1} & \score{17.3}{3.8} & \score{15.2}{3.3}
& \score{12.4}{1.0} & \score{15.3}{3.5} & \score{32.4}{8.0}
& \score{43.4}{6.1} & \bestscore{60.4}{7.3} \\
& hopp-med & med-e
& \score{14.7}{3.6} & \score{15.4}{2.5} & \score{21.1}{9.3}
& \score{14.2}{1.8} & \score{15.1}{3.6} & \score{24.2}{3.6}
& \score{23.4}{6.4} & \bestscore{30.5}{17.2} \\
& hopp-med & expert
& \score{12.5}{1.6} & \score{19.3}{10.5} & \score{12.7}{1.7}
& \score{11.8}{0.9} & \score{14.8}{4.0} & \score{33.7}{7.8}
& \score{37.1}{14.1} & \bestscore{48.2}{20.8} \\
& walk-med & med
& \score{28.1}{12.9} & \score{28.4}{13.7} & \score{38.0}{11.2}
& \score{21.4}{7.0} & \score{22.1}{8.4} & \score{36.6}{2.3}
& \score{41.6}{4.0} & \bestscore{46.5}{5.5} \\
& walk-med & med-e
& \score{35.7}{4.7} & \score{30.7}{9.7} & \score{40.9}{7.2}
& \score{34.0}{9.9} & \score{35.4}{9.1} & \score{44.8}{7.5}
& \score{36.9}{6.5} & \bestscore{48.8}{8.5} \\
& walk-med & expert
& \score{37.3}{8.0} & \score{36.0}{7.0} & \score{41.3}{8.6}
& \score{39.5}{3.8} & \score{36.2}{13.6} & \score{44.0}{4.0}
& \bestscore{49.9}{10.1} & \score{42.6}{7.6} \\
& ant-med & med
& \score{10.2}{1.8} & \score{9.4}{0.9} & \score{12.4}{2.0}
& \score{11.7}{1.0} & \score{11.3}{1.3} & \bestscore{45.1}{12.4}
& \score{42.6}{8.4} & \score{40.8}{10.3} \\
& ant-med & med-e
& \score{9.4}{1.2} & \score{10.0}{0.9} & \score{11.6}{1.3}
& \score{10.2}{1.2} & \score{9.4}{1.4} & \bestscore{33.9}{5.4}
& \score{23.2}{6.1} & \score{25.0}{11.7} \\
& ant-med & expert
& \score{10.2}{0.3} & \score{9.8}{0.6} & \score{11.8}{0.4}
& \score{9.5}{0.6} & \score{9.7}{1.6} & \score{33.2}{9.0}
& \bestscore{35.6}{5.6} & \score{15.9}{4.7} \\
\midrule
Morph & half-med & med
& \score{24.5}{2.4} & \score{21.0}{3.9} & \score{24.2}{5.6}
& \score{18.1}{1.8} & \score{23.7}{3.4} & \score{21.1}{7.6}
& \score{19.5}{2.2} & \bestscore{57.9}{4.4} \\
& half-med & expert
& \score{11.1}{1.5} & \score{11.3}{0.6} & \score{9.1}{1.7}
& \score{10.1}{1.6} & \score{10.1}{0.8} & \score{7.4}{2.3}
& \score{7.7}{1.2} & \bestscore{12.3}{1.6} \\
& hopp-med & med
& \score{15.9}{6.8} & \score{17.8}{10.1} & \score{12.8}{0.1}
& \score{21.7}{7.7} & \score{25.3}{9.7} & \score{16.4}{7.1}
& \score{43.1}{23.9} & \bestscore{69.4}{13.2} \\
& hopp-med & expert
& \score{15.5}{4.1} & \score{13.0}{1.4} & \score{4.5}{4.0}
& \score{10.4}{3.1} & \score{13.4}{2.0} & \score{14.0}{4.0}
& \score{53.2}{50.3} & \bestscore{62.4}{40.9} \\
& walk-med & med
& \score{31.5}{8.6} & \score{35.0}{10.8} & \score{26.7}{6.6}
& \score{38.6}{5.1} & \score{38.5}{8.4} & \score{42.5}{3.1}
& \bestscore{56.7}{8.1} & \score{52.8}{8.2} \\
& walk-med & expert
& \score{37.0}{6.0} & \score{43.7}{5.5} & \score{31.3}{8.6}
& \score{43.8}{8.4} & \score{38.5}{4.3} & \bestscore{49.9}{5.4}
& \score{32.9}{7.3} & \score{33.2}{18.4} \\
& ant-med & med
& \score{71.4}{2.4} & \score{71.5}{6.8} & \score{54.8}{13.2}
& \score{72.8}{2.2} & \score{71.8}{2.7} & \score{75.1}{2.3}
& \bestscore{77.2}{2.8} & \score{76.2}{2.4} \\
& ant-med & expert
& \score{63.9}{8.1} & \score{68.9}{7.6} & \score{47.5}{11.9}
& \score{59.0}{4.7} & \score{66.1}{4.0} & \score{63.8}{6.4}
& \score{74.1}{21.2} & \bestscore{79.0}{12.9} \\
\midrule
\multicolumn{3}{l}{Total (displayed means)}
& 561.7 & 580.2 & 535.7 & 556.2 & 571.6 & 723.5 & 820.9 & \textbf{940.9} \\
\bottomrule
\end{tabular}%
}
\end{table}

\subsection{Detailed Ablations}
\label{app:detailed-ablations}

We provide detailed results for reward quantiles, VAE design,
source-field controls, conditioning horizons, target-data size and
sampling, and fixed mixing weights. Unless otherwise stated, the
experiments cover eight medium-to-medium gravity and morphology
settings and follow the main evaluation protocol.

\subsubsection{Reward Quantile}
\label{app:reward-quantile}

Table~\ref{tab:ablation-quantile} compares reward quantiles.
Our default $q=0.2$ achieves the highest aggregate score,
although the best value varies by task.

\begin{table}[!htbp]
\centering
\caption{Comparison of reward quantiles across eight medium-to-medium
gravity and morphology settings. Values report mean $\pm$ standard
deviation over five seeds, with the highest mean bolded.}
\label{tab:ablation-quantile}
\small
\setlength{\tabcolsep}{4pt}
\resizebox{\textwidth}{!}{%
\begin{tabular}{llcccccc}
\toprule
& & \multicolumn{6}{c}{Reward quantile $q$} \\
\cmidrule(lr){3-8}
Shift & Task & \shortstack{0 (no augmentation)}
& 0.1 & 0.2 (ours) & 0.3 & 0.4 & 0.5 \\
\midrule
Gravity & Ant
& \score{29.8}{4.9} & \score{34.5}{11.9}
& \bestscore{40.8}{10.3} & \score{36.8}{14.1}
& \score{37.8}{4.4} & \score{31.5}{6.9} \\
& HalfCheetah
& \score{42.0}{2.3} & \score{49.4}{2.7}
& \score{47.3}{6.6} & \score{49.5}{3.6}
& \score{48.3}{3.9} & \bestscore{51.4}{2.6} \\
& Hopper
& \score{52.7}{12.5} & \bestscore{62.0}{14.0}
& \score{60.4}{7.3} & \score{60.2}{8.6}
& \score{58.7}{6.5} & \score{41.9}{12.9} \\
& Walker2d
& \score{30.6}{4.3} & \score{34.7}{3.2}
& \bestscore{46.5}{5.5} & \score{42.4}{7.9}
& \score{41.4}{10.0} & \score{40.8}{10.9} \\
\midrule
Morphology & Ant
& \score{76.9}{0.3} & \score{77.1}{0.2}
& \score{76.2}{2.4} & \bestscore{78.0}{0.3}
& \bestscore{78.0}{0.1} & \score{76.8}{1.3} \\
& HalfCheetah
& \score{22.8}{2.4} & \score{54.5}{5.0}
& \score{57.9}{4.4} & \bestscore{58.0}{3.8}
& \score{51.9}{2.3} & \score{50.9}{6.1} \\
& Hopper
& \score{55.5}{6.4} & \score{67.7}{8.2}
& \bestscore{69.4}{13.2} & \score{53.3}{16.4}
& \score{65.7}{14.6} & \score{66.3}{10.9} \\
& Walker2d
& \score{49.2}{5.1} & \score{42.9}{15.7}
& \bestscore{52.8}{8.2} & \score{47.3}{18.6}
& \score{51.3}{8.6} & \score{47.1}{4.6} \\
\midrule
\multicolumn{2}{l}{Total (displayed means)}
& 359.5 & 422.8 & \textbf{451.3} & 425.5 & 433.1 & 406.7 \\
\bottomrule
\end{tabular}%
}
\end{table}

\subsubsection{VAE Design}
\label{app:vae-design}

Table~\ref{tab:ablation-vae} compares uniformly pooled VAE training
(base), balanced source--target sampling (bal), and balanced sampling
with reward regularization (bal + reg). Balanced sampling improves the
aggregate score from $394.6$ to $432.7$, and reward regularization
further increases it to $451.3$. The effects vary by task.

\begin{table}[!htbp]
\centering
\caption{Comparison of VAE design across eight medium-to-medium
gravity and morphology settings. Values report mean $\pm$ standard
deviation over five seeds, with the highest mean bolded.}
\label{tab:ablation-vae}
\small
\begin{adjustbox}{max width=0.6\textwidth}
\setlength{\tabcolsep}{6pt}
\begin{tabular}{llccc}
\toprule
& & \multicolumn{3}{c}{Shared VAE} \\
\cmidrule(lr){3-5}
Shift & Task & base & bal & bal + reg (ours) \\
\midrule
Gravity & Ant
& \score{35.8}{6.2} & \score{31.5}{13.9}
& \bestscore{40.8}{10.3} \\
& HalfCheetah
& \bestscore{48.3}{5.0} & \score{47.6}{4.5}
& \score{47.3}{6.6} \\
& Hopper
& \score{39.2}{19.6} & \bestscore{60.4}{7.8}
& \bestscore{60.4}{7.3} \\
& Walker2d
& \score{37.3}{6.5} & \score{42.1}{8.2}
& \bestscore{46.5}{5.5} \\
\midrule
Morphology & Ant
& \bestscore{78.1}{0.2} & \score{77.9}{0.2}
& \score{76.2}{2.4} \\
& HalfCheetah
& \bestscore{58.3}{2.8} & \score{57.3}{5.1}
& \score{57.9}{4.4} \\
& Hopper
& \score{64.0}{10.7} & \score{68.2}{12.0}
& \bestscore{69.4}{13.2} \\
& Walker2d
& \score{33.6}{12.3} & \score{47.7}{9.5}
& \bestscore{52.8}{8.2} \\
\midrule
\multicolumn{2}{l}{Total (displayed means)}
& 394.6 & 432.7 & \textbf{451.3} \\
\bottomrule
\end{tabular}
\end{adjustbox}
\end{table}

\subsubsection{Source-Field and High-Reward Resampling Controls}
\label{app:mechanism-controls}

We compare two controls while keeping the VAE and normalization fixed.
The target-only field learns the low-to-high field from target
transitions and applies it to the same target anchors.
High-reward resampling repeats real high-reward target transitions
with replacement. Each uses a 1,000-transition augmentation budget;
RIFT retains up to 1,000 synthetic transitions after filtering.

\begin{table}[!htbp]
\centering
\caption{Source-field and resampling controls. Values report
mean $\pm$ standard deviation. IQL$^*$ and RIFT use the original
paper reference results.}
\label{tab:mechanism-controls}
\small
\setlength{\tabcolsep}{4pt}
\begin{adjustbox}{max width=\textwidth}
\begin{tabular}{llrrrr}
\toprule
Shift & Task & IQL$^*$ & High-reward resampling
& Target-only field & RIFT \\
\midrule
Gravity & Ant
& $29.8\pm4.9$ & $36.1\pm6.7$ & $27.3\pm9.7$ & $40.8\pm10.3$ \\
& HalfCheetah
& $42.0\pm2.3$ & $47.3\pm7.1$ & $48.4\pm4.4$ & $47.3\pm6.6$ \\
& Hopper
& $52.7\pm12.5$ & $52.2\pm8.5$ & $61.1\pm10.2$ & $60.4\pm7.3$ \\
& Walker2d
& $30.6\pm4.3$ & $42.5\pm6.7$ & $38.3\pm7.9$ & $46.5\pm5.5$ \\
\midrule
Morphology & Ant
& $76.9\pm0.3$ & $77.0\pm0.4$ & $77.6\pm0.5$ & $76.2\pm2.4$ \\
& HalfCheetah
& $22.8\pm2.4$ & $19.2\pm2.4$ & $56.7\pm5.1$ & $57.9\pm4.4$ \\
& Hopper
& $55.5\pm6.4$ & $53.2\pm7.1$ & $63.5\pm6.7$ & $69.4\pm13.2$ \\
& Walker2d
& $49.2\pm5.1$ & $57.0\pm7.0$ & $50.9\pm8.9$ & $52.8\pm8.2$ \\
\midrule
\multicolumn{2}{l}{Total (displayed means)}
& 359.5 & 384.5 & 423.8 & \textbf{451.3} \\
\bottomrule
\end{tabular}
\end{adjustbox}
\end{table}

RIFT exceeds the target-only field by $27.5$ aggregate points and
high-reward resampling by $66.8$ points, although the advantage varies
by task. The target-only field retains source information in the shared
VAE and tests where the field is learned, rather than providing a fully
source-free control. These comparisons do not directly measure field
invariance or isolate OT from alternative source couplings.

\subsubsection{Horizon of the Reward-Class Signal}
\label{app:reward-horizon}

Immediate reward describes the transition being augmented, while longer-horizon returns also depend on subsequent actions and future dynamics. We compare immediate-reward classes with classes defined by the summed reward over the next $n$ steps within each transition's
episode, for $n\in\{5,10,20\}$.

The aggregate scores change by $+0.2\%$, $-2.9\%$, and $-0.1\%$ relative to immediate-reward conditioning. We hypothesize that dense rewards already provide useful local task information, limiting the benefit of longer-horizon labels. These results support immediate reward as a simple, effective choice in the evaluated tasks.

\begin{table}[!htbp]
\centering
\caption{Conditioning-horizon comparison across eight medium-to-medium
gravity and morphology settings. Values report mean $\pm$ standard
deviation. The one-step column uses the original paper results.}
\label{tab:reward-horizon}
\small
\begin{adjustbox}{max width=\textwidth}
\begin{tabular}{llrrrr}
\toprule
Shift & Task & 1 step & 5 steps & 10 steps & 20 steps \\
\midrule
Gravity & Ant
& $40.8\pm10.3$ & $35.9\pm2.2$ & $44.9\pm8.9$ & $40.3\pm9.4$ \\
& HalfCheetah
& $47.3\pm6.6$ & $48.3\pm2.9$ & $51.3\pm2.1$ & $50.7\pm2.1$ \\
& Hopper
& $60.4\pm7.3$ & $54.4\pm6.8$ & $53.2\pm12.5$ & $60.5\pm6.6$ \\
& Walker2d
& $46.5\pm5.5$ & $46.1\pm3.7$ & $36.5\pm4.6$ & $39.9\pm11.3$ \\
\midrule
Morphology & Ant
& $76.2\pm2.4$ & $76.6\pm2.8$ & $76.3\pm1.8$ & $74.2\pm1.1$ \\
& HalfCheetah
& $57.9\pm4.4$ & $59.0\pm2.1$ & $57.2\pm2.8$ & $57.2\pm4.4$ \\
& Hopper
& $69.4\pm13.2$ & $77.7\pm5.3$ & $62.6\pm7.5$ & $79.4\pm4.1$ \\
& Walker2d
& $52.8\pm8.2$ & $54.0\pm9.5$ & $56.1\pm10.6$ & $48.8\pm5.8$ \\
\midrule
\multicolumn{2}{l}{Total (displayed means)}
& 451.3 & 452.0 & 438.1 & 451.0 \\
\bottomrule
\end{tabular}
\end{adjustbox}
\end{table}

\subsubsection{Sensitivity to Target-Data Size}
\label{app:target-size}

We evaluate 500, 1,000, 2,500, and 5,000 target transitions.
The 5,000-transition setting uses the original paper results.

\begin{table}[!htbp]
\centering
\caption{Aggregate performance at different target-data sizes.
Relative differences use IQL$^*$ as the denominator.}
\label{tab:target-size}
\begin{tabular}{rrrr}
\toprule
Target transitions & IQL$^*$ & RIFT & Relative difference \\
\midrule
500   & 257.6 & 245.6 & $-4.7\%$ \\
1,000 & 288.9 & 293.6 & $+1.6\%$ \\
2,500 & 336.8 & 398.6 & $+18.3\%$ \\
5,000 & 359.5 & 451.3 & $+25.5\%$ \\
\bottomrule
\end{tabular}
\end{table}

RIFT improves aggregate performance at three of the four sizes,
but underperforms at 500 transitions. Source data therefore do not
remove the need for sufficient target information. Limited coverage
or representation quality may contribute to this result, but the
experiment does not isolate the cause.

\subsubsection{Sensitivity to Target-Data Sampling}
\label{app:target-subsets}

Training seeds on a fixed dataset measure training variability,
but not sensitivity to which target transitions are available.
From each 5,000-transition ODRL target dataset, we draw four random
subsets of 4,000 transitions. For each subset, we refit the VAE,
recompute the OT couplings and field, and retrain the policy.

\begin{table}[!htbp]
\centering
\caption{Sensitivity to target-data removal. The full pipeline is
rebuilt for each of four 4,000-transition subsets of the original
5,000-transition target dataset.}
\label{tab:target_subsets}
\small
\begin{tabular}{llrr}
\toprule
Shift & Task & IQL$^*$ & RIFT \\
\midrule
Gravity & Ant
& $22.8\pm5.5$ & $43.5\pm5.6$ \\
& HalfCheetah
& $41.2\pm2.6$ & $48.3\pm2.1$ \\
& Hopper
& $49.2\pm10.8$ & $56.4\pm9.2$ \\
& Walker2d
& $47.8\pm4.4$ & $37.4\pm12.6$ \\
\midrule
Morphology & Ant
& $76.6\pm0.4$ & $77.3\pm0.5$ \\
& HalfCheetah
& $17.2\pm2.5$ & $56.2\pm6.0$ \\
& Hopper
& $49.8\pm19.9$ & $68.8\pm6.1$ \\
& Walker2d
& $48.9\pm9.2$ & $41.1\pm11.1$ \\
\midrule
\multicolumn{2}{l}{Total (displayed means)}
& 353.5 & \textbf{429.0} \\
\bottomrule
\end{tabular}
\end{table}

RIFT achieves higher mean scores in six of eight settings, with
aggregate scores of $429.0$ versus $353.5$. Its advantage therefore
persists after removing some target transitions. The subsets overlap
and come from the same dataset, so this does not establish robustness
across independently collected target datasets.

\subsubsection{Fixed Mixing Weights}
\label{app:fixed-mixing}

For RIFT and the augmentation comparisons, we select the better mixing arm for each task using mean evaluation return across seeds. Table~\ref{tab:fixed-mixing} fixes $\lambda$ across tasks.
\begin{table}[!htbp]
\centering
\caption{Aggregate performance with a fixed mixing weight.
At $\lambda=0.5$, half the policy minibatch uses raw source data;
at $\lambda=1$, none does.}
\label{tab:fixed-mixing}
\begin{tabular}{lrr}
\toprule
Method & $\lambda=0.5$ & $\lambda=1$ \\
\midrule
IQL   & 234.0 & 312.4 \\
SMOTE & 262.1 & 337.2 \\
RIFT  & \textbf{280.5} & \textbf{392.3} \\
\bottomrule
\end{tabular}
\end{table}

RIFT retains the highest aggregate score among these methods under either mixing choice. 

\subsubsection{HalfCheetah Contribution}
Excluding Morph-HalfCheetah from the selected-arm comparison leaves
aggregate scores of $393.4$ for RIFT and $336.7$ for IQL$^*$,
a gain of $56.7$ points. Excluding both HalfCheetah settings leaves
$346.1$ versus $294.7$, a gain of $51.4$ points. The aggregate
advantage is therefore not confined to HalfCheetah.

\subsubsection{Learning Curves}

\begin{figure}[!htbp]
    \centering
    \includegraphics[width=0.6\linewidth]{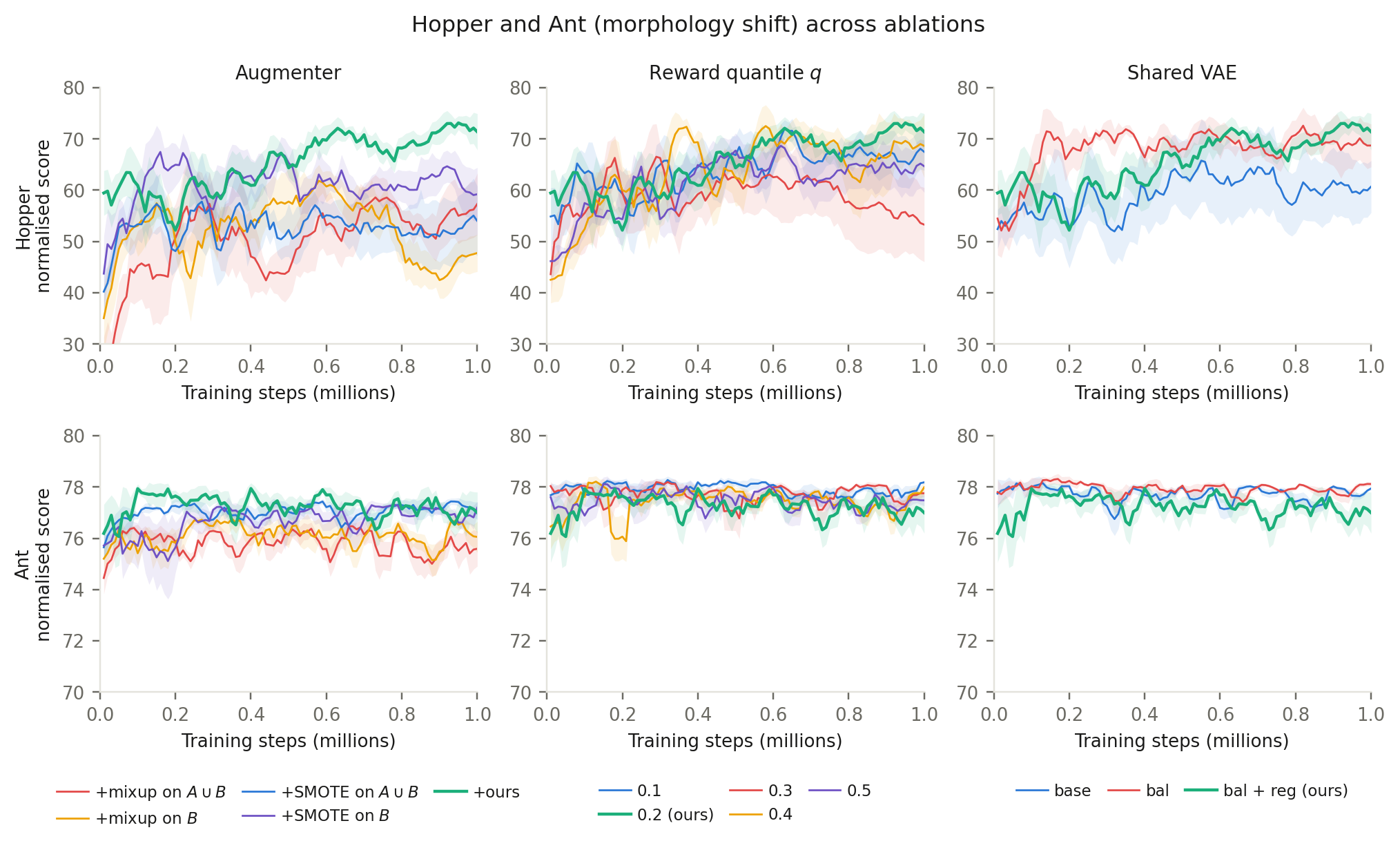}
    \caption{Learning curves for augmentation, reward-quantile, and
    VAE-design ablations under morphology shifts.}
    \label{fig:learning-curves}
\end{figure}

Figure~\ref{fig:learning-curves} shows learning curves for augmentation,
reward-quantile, and VAE-design ablations under morphology shifts.
Differences are larger on Hopper than on Ant.

\subsection{Held-Out Reconstruction and Synthetic-Transition Validity}
\label{app:synthetic-validity}

We filter terminal initial states, and also evaluate both held-out reconstruction and target-simulator replay on synthetic transitions. For these diagnostics, we exclude 500 target transitions (10\%) and 5,000 source transitions from VAE fitting. 

On held-out target transitions, per-dimension reconstruction MSE ranges from 0.001 to 0.07 times the corresponding dimension's variance across the eight settings. We also encode and decode each held-out transition, restore the decoded state, apply the decoded action, and compare the simulator's successor and reward with the generated values.

\paragraph{Generated-transition replay.}
We apply the same replay diagnostic to 1,000 RIFT proposals per setting
and to Mixup and SMOTE proposals built on the same anchors. Successor
error is the norm of the difference between the generated and replayed
successor, divided by the median norm of a real one-step state change
in the corresponding target dataset. An error of 1 therefore represents
a typical real-step displacement. Reward MAE compares the reward stored
in the generated row with the simulator reward and is divided by the
target high-minus-low class mean reward. Realized reward is
$(r_{\mathrm{sim}}-\bar r_{B,\mathrm{low}})/
(\bar r_{B,\mathrm{high}}-\bar r_{B,\mathrm{low}})$;
0 and 1 denote the low- and high-class means. Values can exceed 1.
Table~\ref{tab:synthetic-validity} reports RIFT's results.

\begin{table}[t]
\centering
\caption{Target-simulator replay diagnostics.}
\label{tab:synthetic-validity}
\small
\setlength{\tabcolsep}{4pt}
\resizebox{\textwidth}{!}{%
\begin{tabular}{llrrrr}
\toprule
Shift & Task & \shortstack{Held-out decoded\\successor error}
& \shortstack{Generated\\successor error}
& \shortstack{Generated reward\\MAE (class gaps)}
& \shortstack{Generated\\realized reward}\\
\midrule
Gravity & Ant & 0.25 & 0.57 & 0.18 & 0.87 \\
& HalfCheetah & 0.05 & 0.11 & 0.12 & 0.99 \\
& Hopper & 0.33 & 1.46 & 0.05 & 1.06 \\
& Walker2d & 0.65 & 7.07 & 0.08 & 0.80 \\
\midrule
Morphology & Ant & 0.16 & 0.29 & 0.14 & 0.91\\
& HalfCheetah & 0.05 & 0.08 & 0.05 & 1.08\\
& Hopper & 0.19 & 0.60 & 0.19 & 0.87 \\
& Walker2d & 0.59 & 2.20 & 0.05 & 0.88\\
\bottomrule
\end{tabular}%
}
\end{table}

Replaying real target transitions yields next-step error of at most
0.05 real steps and reward error below 0.003 class gaps. Averaged
equally across the eight settings, RIFT proposals achieve realized
reward of 0.93, compared with 0.57 for Mixup and 0.55 for SMOTE. However, we note that next-step accuracy does not predict downstream gains and that these measurements do not guarantee exactly accurate transitions.
\paragraph{Reconstruction across VAE variants.}
Table~\ref{tab:vae-recon} reports reconstruction variance explained for
HalfCheetah and Walker2d under gravity shifts. This diagnostic is
separate from the held-out per-dimension reconstruction errors above.

\begin{table}[!htbp]
\centering
\caption{Reconstruction variance explained (\%) by the three
shared-VAE variants under gravity shifts. Results are
mean $\pm$ standard deviation.}
\label{tab:vae-recon}
\footnotesize
\setlength{\tabcolsep}{5pt}
\begin{tabular}{llcc}
\toprule
Task & VAE & Source & Target \\
\midrule
HalfCheetah
& base      & $99.17 \pm 0.03$ & $96.41 \pm 0.23$ \\
& bal       & $98.77 \pm 0.04$ & $99.38 \pm 0.06$ \\
& bal + reg & $98.52 \pm 0.03$ & $99.30 \pm 0.06$ \\
\midrule
Walker2d
& base      & $99.36 \pm 0.03$ & $98.32 \pm 0.09$ \\
& bal       & $98.95 \pm 0.02$ & $99.65 \pm 0.03$ \\
& bal + reg & $98.80 \pm 0.02$ & $99.62 \pm 0.03$ \\
\bottomrule
\end{tabular}
\end{table}

\subsection{Friction Results}
Table~\ref{tab:friction} reports results under friction shifts. RIFT obtains the highest aggregate score across the twelve settings, although STC or other baselines perform better on several individual tasks.

\begin{table}
\centering
\caption{Performance under friction shifts when the source dataset is fixed to medium quality and target-data quality varies. Values report the mean $\pm$ standard deviation over five seeds, with the highest mean bolded.}
\label{tab:friction}
\setlength{\tabcolsep}{3pt}
\resizebox{\textwidth}{!}{%
\begin{tabular}{ll*{8}{c}}
\toprule
Source & Target
& IQL & DARA & BOSA & SRPO & IGDF & OTDF & STC
& \textbf{RIFT (ours)} \\
\midrule
halfcheetah-med & med
& \score{69.7}{1.3} & \score{66.2}{4.1} & \score{68.5}{1.2}
& \score{66.7}{2.1} & \score{66.1}{1.9} & \score{66.8}{0.6}
& \score{67.4}{1.0} & \bestscore{72.4}{2.6} \\
halfcheetah-med & med-e
& \score{66.8}{0.6} & \score{60.4}{11.9} & \score{69.2}{0.6}
& \score{66.8}{3.9} & \score{68.6}{0.6} & \score{61.3}{5.4}
& \score{69.4}{1.0} & \bestscore{71.4}{1.4} \\
halfcheetah-med & expert
& \score{68.2}{0.4} & \score{67.9}{0.5} & \score{69.7}{1.3}
& \score{67.1}{2.3} & \score{68.2}{2.5} & \score{68.3}{0.6}
& \score{67.3}{0.6} & \bestscore{70.5}{3.2} \\
hopper-med & med
& \score{24.5}{1.7} & \score{24.2}{2.2} & \score{25.4}{1.9}
& \score{26.5}{2.0} & \score{27.2}{4.3} & \score{29.3}{4.6}
& \score{27.4}{2.0} & \bestscore{38.8}{2.9} \\
hopper-med & med-e
& \bestscore{26.4}{2.5} & \score{21.7}{5.6} & \score{23.0}{1.7}
& \score{24.3}{3.7} & \score{25.7}{2.3} & \bestscore{26.4}{2.4}
& \score{23.4}{5.1} & \score{23.8}{11.0} \\
hopper-med & expert
& \score{21.5}{1.1} & \score{24.6}{2.5} & \score{25.7}{0.8}
& \score{21.1}{2.7} & \score{23.6}{3.8} & \score{25.9}{1.1}
& \bestscore{41.6}{26.1} & \score{37.8}{15.2} \\
walker2d-med & med
& \score{72.6}{6.2} & \score{73.9}{11.7} & \score{72.1}{5.2}
& \score{67.5}{6.0} & \score{70.3}{3.6} & \score{70.4}{6.6}
& \score{73.6}{8.6} & \bestscore{74.1}{6.2} \\
walker2d-med & med-e
& \score{59.0}{1.3} & \score{61.5}{20.2} & \score{42.8}{17.5}
& \score{57.4}{7.9} & \score{59.1}{5.4} & \score{59.9}{10.2}
& \bestscore{71.2}{5.8} & \score{69.7}{10.0} \\
walker2d-med & expert
& \score{52.5}{3.6} & \score{60.0}{11.7} & \score{51.5}{12.9}
& \score{52.3}{12.7} & \score{54.9}{6.1} & \score{49.6}{14.6}
& \bestscore{67.9}{6.6} & \score{64.1}{11.9} \\
ant-med & med
& \score{58.2}{2.3} & \score{58.7}{2.0} & \score{57.6}{4.0}
& \score{56.9}{2.5} & \score{55.2}{3.2} & \score{58.3}{0.2}
& \bestscore{60.2}{3.1} & \score{58.5}{0.6} \\
ant-med & med-e
& \score{59.3}{2.5} & \score{58.3}{1.0} & \bestscore{60.5}{0.3}
& \score{58.4}{0.4} & \score{57.8}{0.5} & \score{58.1}{0.4}
& \score{45.7}{10.5} & \score{58.6}{0.4} \\
ant-med & expert
& \score{58.2}{0.3} & \score{58.3}{0.5} & \score{59.3}{1.8}
& \score{57.2}{2.0} & \score{58.3}{0.2} & \score{56.5}{2.3}
& \score{60.6}{1.8} & \bestscore{66.0}{16.7} \\
\midrule
\multicolumn{2}{l}{Total (displayed means)}
& 636.9 & 635.7 & 625.3 & 622.2 & 635.0 & 630.8 & 675.7 & \textbf{705.7} \\
\bottomrule
\end{tabular}%
}
\end{table}

\subsection{Comparison with DmC}
\label{app:DMC-results}
We additionally compare against DmC~\citep{van2025dmc}, whose official implementation is available in \citep{dmcgithub}, on the gravity-shift settings reported in its paper. Table~\ref{tab:dmc-source-quality} reports best-over-training scores: the maximum normalized score over evaluation checkpoints for each seed, followed by the mean and population standard deviation across seeds. This statistic selects checkpoints using target returns, unlike the final-checkpoint statistic in the main experiments, so we present this comparison separately.

\begin{table}
\centering
\caption{Best-over-training performance under gravity shifts when the target dataset is fixed to medium quality and source-data quality varies. Values report the mean $\pm$ population standard deviation over five seeds, with the highest mean bolded.}
\label{tab:dmc-source-quality}
\begin{tabular}{llcc}
\toprule
Source & Target & DmC & \textbf{RIFT (ours)} \\
\midrule
half-med   & med & \score{48.0}{0.6} & \bestscore{55.0}{1.0} \\
half-med-r & med & \score{32.2}{0.8} & \bestscore{43.6}{1.9} \\
half-med-e & med & \score{49.9}{1.4} & \bestscore{56.6}{0.6} \\
\midrule
hopp-med   & med & \score{30.6}{5.5} & \bestscore{74.4}{5.2} \\
hopp-med-r & med & \score{35.4}{0.7} & \bestscore{71.9}{4.3} \\
hopp-med-e & med & \score{52.2}{6.6} & \bestscore{75.0}{2.1} \\
\midrule
walk-med   & med & \score{52.5}{2.0} & \bestscore{56.2}{3.0} \\
walk-med-r & med & \score{42.2}{5.9} & \bestscore{53.9}{4.3} \\
walk-med-e & med & \score{57.5}{3.3} & \bestscore{68.0}{5.9} \\
\midrule
ant-med    & med & \score{56.9}{2.2} & \bestscore{64.0}{5.6} \\
ant-med-r  & med & \score{41.8}{4.7} & \bestscore{67.8}{4.3} \\
ant-med-e  & med & \score{55.7}{8.1} & \bestscore{59.9}{4.0} \\
\midrule
\multicolumn{2}{l}{Total (displayed means)}
& 554.9 & \textbf{746.3} \\
\bottomrule
\end{tabular}
\end{table}

\subsection{Computing Infrastructure}

We list the computing infrastructure used to run all experiments in this paper in Table~\ref{tab:compute}. All experiments are CPU-only. Each IQL training run took around 2 hours on a training core, the main bottleneck for compute.

\begin{table}[!htbp]
\centering
\caption{Computing infrastructure used to run all experiments in this paper.}
\label{tab:compute}
\begin{tabular}{ll}
\toprule
Component & Specification \\
\midrule
CPU & AMD EPYC 9474F / 9654 \\
Cores & up to 96 in parallel \\
Memory & 4\,GB per run \\
\bottomrule
\end{tabular}
\end{table}

\end{document}